\documentclass{article}
\usepackage{natbib}

\usepackage{arxiv}
\usepackage[utf8]{inputenc} % allow utf-8 input
\usepackage[T1]{fontenc}    % use 8-bit T1 fonts
\usepackage{hyperref}       % hyperlinks
\usepackage{url}            % simple URL typesetting
\usepackage{booktabs}       % professional-quality tables
\usepackage{amsfonts}       % blackboard math symbols
\usepackage{nicefrac}       % compact symbols for 1/2, etc.
\usepackage{microtype}      % microtypography
\usepackage{lipsum}		% Can be removed after putting your text content
\usepackage{graphicx}
\usepackage{doi}
\usepackage{xcolor}
\usepackage{svg}
\usepackage{subcaption} % The recommended package for subfigures
\usepackage{booktabs}
\usepackage{siunitx}
\usepackage{geometry}
\usepackage{graphicx}
\usepackage{arxiv}
\usepackage{subcaption}  % In the preamble
\usepackage{graphicx}% Include figure files
\usepackage{dcolumn}% Align table columns on decimal point
\usepackage{bm}% bold math
\usepackage{amsmath}
\usepackage{algorithm}
\usepackage{algorithmic}
\usepackage{subcaption}
\usepackage{soul}
\usepackage{color}
\usepackage{xcolor}

\title{Adaptive Quantum Physics-Informed Neural Networks for Differential Equations with Applications to Fluid Dynamics}

\author{ 
\href{https://orcid.org/0000-0002-9650-3619
}{\includegraphics[scale=0.06]{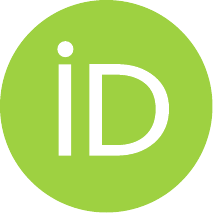}
\hspace{2mm}Fábio Pereira dos Santos}\\
Coordination of Mathematical and Computational Methods\\
National Laboratory of Scientific Computing\\
Av. Getúlio Vargas, 333 - Quitandinha, Petrópolis - RJ, 25651-075 \\
\texttt{fpsantos@lncc.br} \\
\And
\href{https://orcid.org/0000-0001-5472-3544
}{\includegraphics[scale=0.06]{orcid.pdf}
\hspace{2mm}Renato Portugal}\\
Coordination of Mathematical and Computational Methods\\
National Laboratory of Scientific Computing\\
Av. Getúlio Vargas, 333 - Quitandinha, Petrópolis - RJ, 25651-075 \\
\texttt{portugal@lncc.br} \\
\And
\href{https://orcid.org/0000-0001-5472-3544
}{\includegraphics[scale=0.06]{orcid.pdf}
\hspace{2mm}Júlio de Castro Vargas Fernandes}\\
Coordination of Mathematical and Computational Methods\\
National Laboratory of Scientific Computing\\
Av. Getúlio Vargas, 333 - Quitandinha, Petrópolis - RJ, 25651-075 \\
\texttt{julio@lncc.br} \\
\And
\href{https://orcid.org/0000-0002-2941-2522
}{\includegraphics[scale=0.06]{orcid.pdf}
\hspace{2mm}Lucas Timotheo Sanches}\\
Center for Computation and Technology - Louisiana State University,\\ 888 S. Stadium Dr., Baton Rouge, LA 70808, USA
\texttt{lsanches@lsu.edu} \\
}

\renewcommand{\shorttitle}{\textit{Adaptive Quantum Physics-Informed Neural Networks}}

\begin{document}
\maketitle

\begin{abstract}
 Physics-informed neural networks (PINNs) have emerged as a versatile approach for solving nonlinear partial differential equations (PDEs), yet achieving high accuracy efficiently using these techniques remains challenging for high-dimensional or multiscale systems. Here, we present a hybrid quantum-classical framework that enhances Quantum PINNs (QPINNs) through adaptive collocation point sampling and loss-aware attention mechanisms. By dynamically prioritizing points in regions with large PDE residuals or steep solution gradients, our method mitigates the spectral bias inherent in conventional PINNs. Current Quantum Physics-Informed Neural Networks are commonly assumed to be limited by the expressive power of quantum circuits. In our work, we observed that, across diverse differential equations, optimization — not only expressivity — can be an important bottleneck. Furthermore, a trainable loss-weighting scheme balances contributions from physics residuals, boundary conditions, and data fidelity during training. Integrating these strategies with quantum computing techniques (including variational quantum circuits and quantum gradient estimation) can yield at least a 60\% improvement in solution accuracy under specific regimes for benchmark fluid flows and reaction-diffusion systems. Finally, we argue that merely increasing model expressivity is insufficient for resolving complex PDEs via QPINNs, as they remain constrained by the structural optimization limitations of classical PINNs. This framework provides a scalable pathway for quantum-enhanced scientific machine learning, bridging physics-based modeling with emerging quantum computational capabilities.
\end{abstract}

\keywords{Physics-informed neural networks, Quantum Computing, Quantum Neural Network, Partial Differential Equations}%Use showkeys class option if keyword
                              %display desired
\section{\label{sec:Intro}Introduction}

Turbulence remains one of the most challenging phenomena in classical physics, famously described by Feynman as “the most important unsolved problem of classical physics”~\citep{Feynman1964}. Its analysis requires solving nonlinear partial differential equations (PDEs) that govern fluid dynamics across a wide range of spatial and temporal scales, posing substantial computational challenges. Beyond fluid mechanics, PDEs are central to diverse fields such as climate modeling, magnetohydrodynamics, and astrophysics~\cite{Gaitan2021}. However, high-resolution simulations of such systems often exceed the capabilities of classical computational approaches, motivating the exploration of alternative paradigms, including quantum computing~\cite{Gaitan2021,Yepez1999}.

Quantum computing (QC) has emerged as a rapidly advancing framework with the potential to accelerate specific classes of computational problems. Although classical methods are expected to remain dominant for many applications in the near term, even limited quantum advantages in PDE solvers could yield significant scientific and technological impact. This raises fundamental open questions: which classes of PDEs are most amenable to quantum acceleration, and to what extent can QC contribute to the solution of complex multiscale systems, such as the Navier–Stokes equations? Addressing these questions remains an active and largely unexplored area of research.

We are currently in the noisy intermediate-scale quantum (NISQ) era, characterized by quantum processors with a few hundred qubits~\cite{Preskill2018,Yepez1999}. As a rough illustration, a reactive system discretized with approximately 1 billion mesh points could theoretically be represented with about 30 qubits~\cite{Neau2020}. However, classical algorithms often do not map efficiently to quantum hardware, particularly when considering decoherence, noise, and the requirements of fault tolerance~\cite{Childs2018,Yepez1999}. This underscores the need for algorithms specifically tailored to NISQ devices.

Current strategies for quantum PDE solvers can be grouped into three categories: Quantum Fourier Transform (QFT)-based methods~\cite{Steijl2020,Steijl2018}, quantum Lattice-Boltzmann methods (LBM)~\cite{Yepez2002,Kokail2019}, and discretization-based methods~\cite{Oz2022,Berry2014,Montanaro2016,Costa2019}. QFT-based approaches, for instance, solve the Poisson form of the NSE using vortex-in-cell techniques on quantum circuits with up to 24 qubits and $256^3$ degrees of freedom~\cite{Steijl2018}. While these methods are not yet competitive with classical FFT-based solvers, they provide a framework for potential future improvements if coherence and scalability can be enhanced. LBM-based methods simulate fluid dynamics via quantum analogs of kinetic particle models~\cite{Yepez2002}. These approaches operate primarily on quantum hardware, reducing classical–quantum communication, but extracting the full solution still requires extensive measurements. Discretization-based methods reformulate PDEs as linear or nonlinear systems that can be addressed with quantum algorithms, including amplitude estimation~\cite{Brassard2002}, Harrow-Hassidim-Lloyd (HHL)~\cite{Harrow2009}, and Variational Quantum Linear Solver (VQLS) approaches~\cite{BravoPrieto2019}. While HHL offers logarithmic scaling in theory, practical application to large-scale PDEs remains limited. {A promising new formulation, introduced in~\cite{Jin2024Schrodingerization}, transforms the classical PDE into a Schr\"odinger-like form to enable quantum mechanical treatment and potential quantum computational algorithms}.

Despite these advances, two major challenges {remain}: measurement overhead on NISQ devices and the reliance on classical discretization, which constrains potential speedups. To address these limitations, we explore an alternative direction: integrating machine learning, specifically Physics-Informed Neural Networks (PINNs), with quantum computing. PINNs provide a mesh-free framework, {although not mesh independent which makes them amenable to adaptive schemes, as our work proposes}, in which neural networks approximate PDE solutions by embedding physical laws into the loss function. This paradigm can extend to more general differential systems, including integro-differential and fractional PDEs. Although the combination of PINNs and quantum architectures is still in its early stages, it offers a promising avenue for developing quantum-compatible solvers and may inform future strategies for tackling high-dimensional and multiscale PDEs \cite{Berger2025,e26080649,6nh4-yh2y}. {Our work observed that QPINNs can share the same optimization limitations as classical PINNs. {We focus on the hybrid quantum / classical training loop, rather than quantum speedups or specific encoding techniques and illustrate the importance of classical optimization strategies in reducing approximation error in QPINNs}. {Our} analysis suggests that enhancing quantum expressivity should be paired with robust classical training schemes to effectively resolve complex PDEs {via QPINNs}. Previous studies have already demonstrated that hybrid QPINNs can outperform comparable classical PINNs under specific settings \cite{Berger2025,e26080649,6nh4-yh2y}. Therefore, this work does not revisit that comparison. 

\subsection{\label{sec:PINN}PHYSICS-INFORMED NEURAL NETWORKS}

Physics-Informed Neural Networks (PINNs) provide a flexible framework for approximating solutions to partial differential equations (PDEs) by embedding the underlying physical laws, together with initial and boundary conditions, directly into the loss function~\cite{raissi2019physics,Raissi2017PartI,Raissi2017PartII,Cuomo2022}. {Rather than relying on full-field labeled solution datasets, PINNs enforce the PDE residual at sampled collocation points in the spatiotemporal domain, thereby constraining the learned solution using the structure of the governing equations}~\cite{Karniadakis2021}. In PINNs, the solution of a PDE is represented by a neural network that maps input coordinates (e.g., spatial and temporal variables) to the corresponding solution fields or quantities of interest. The training process is formulated as the minimization of a composite loss function, which enforces the PDE residual, boundary and initial conditions, and, when available, observational data. Through this mechanism, PINNs effectively recast the solution of PDEs as a {physics-}constrained optimization problem, enabling learning of solutions that adhere to the prescribed physical laws.

Introduced by \citet{Raissi2017PartI,Raissi2017PartII} and extended by \citet{raissi2019physics}, PINNs were shown to approximate solutions to PDEs such as the Schrödinger and Burgers equations, while also enabling inverse problems that infer unknown parameters from noisy data. Since then, PINNs have found broad application in fluid mechanics, biology, and computational physics~\cite{Karniadakis2021}.

Consider a general partial differential equation (PDE) of the form:
\begin{equation}
\mathcal{F}(\mathbf{u}(\mathbf{z}); \mathbf{q}) = \mathbf{f}(\mathbf{z}), \quad \mathbf{z} \in \Omega,
\qquad
\mathcal{B}(\mathbf{u}(\mathbf{z})) = \mathbf{g}(\mathbf{z}), \quad \mathbf{z} \in \partial \Omega,
\end{equation}
where $\Omega \subset \mathbb{R}^{n+1}$ denotes the space--time domain with boundary $\partial \Omega$, and $\mathbf{z} = [x_1, \dots, x_n, t]$ represents the spatial and temporal coordinates. The unknown field is given by $\mathbf{u} : \Omega \rightarrow \mathbb{R}^m$, while $\mathbf{q}$ denotes a set of physical parameters characterizing the system. The operator $\mathcal{F}$ defines the governing differential equation, and $\mathcal{B}$ encodes the associated boundary (and, when applicable, initial) conditions. The functions $\mathbf{f} : \Omega \rightarrow \mathbb{R}^m$ and $\mathbf{g} : \partial \Omega \rightarrow \mathbb{R}^m$ represent known source and boundary terms, respectively.

Physics-Informed Neural Networks (PINNs) approximate the solution $\mathbf{u}(\mathbf{z})$ of a partial differential equation by a neural network $\mathbf{u}_{\theta}(\mathbf{z})$ parameterized by $\theta$:

\begin{equation}
\mathbf{u}(\mathbf{z}) \approx \mathbf{u}_{\theta}(\mathbf{z}).
\end{equation}

The network parameters are obtained by minimizing a composite loss function:

\begin{equation}
\theta^{*} = \arg\min_{\theta} \left( \omega_{p} \mathcal{L}_p + \omega_{b} \mathcal{L}_b + \omega_{d} \mathcal{L}_d + \omega_{i} \mathcal{L}_i \right),
\end{equation}

where $\mathcal{L}_p$, $\mathcal{L}_b$, $\mathcal{L}_i$ and $\mathcal{L}_d$ enforce the PDE residuals, boundary and initial conditions, and available data, respectively, and the weights $\omega$ balance their relative contributions. Figure~\ref{fig:PINFIG} illustrates the PINN framework. Collocation points $\mathbf{z} \in \Omega$ (in Figure~\ref{fig:PINFIG} represented by x and t) are sampled within the domain, and the network outputs the corresponding approximate solution $\mathbf{u}_{\theta}(\mathbf{z})$. Automatic differentiation is employed to evaluate derivatives required for the PDE residuals, which are then incorporated into the loss function. Training proceeds by optimizing the network parameters to minimize this combined loss, thereby producing a solution that simultaneously satisfies the governing equations, boundary conditions, and any observed data.

\begin{figure}[!ht]
\centering
\includegraphics[width=0.8\textwidth]{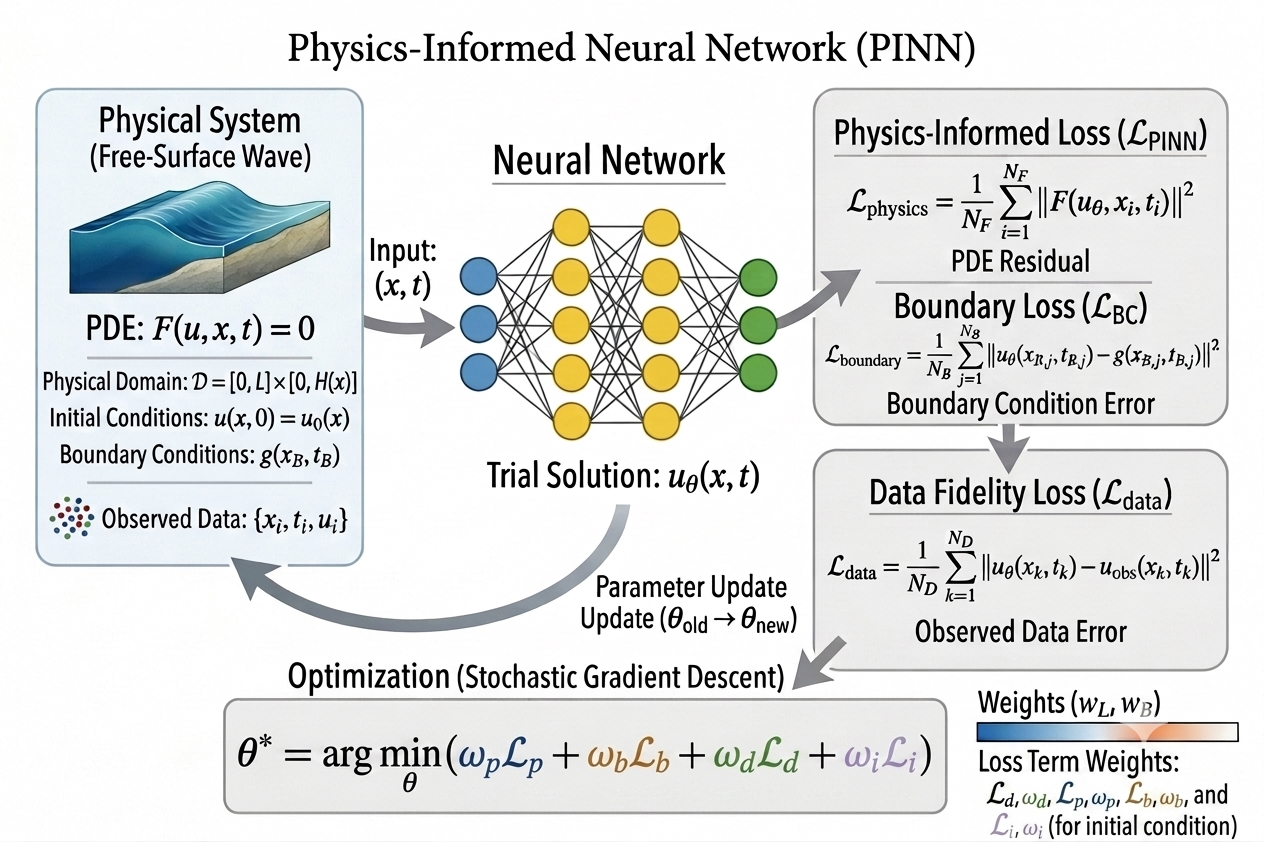}
\caption{Schematic of the physics-informed neural network (PINN) for free-surface wave problems. The network takes $(x, t)$ as input and outputs the trial solution $u_\theta(x, t)$. The loss function  combines data fidelity $\mathcal{L}_{\text{data}}$, PDE residual $\mathcal{L}_{\text{physics}}$, and boundary condition $\mathcal{L}_{\text{boundary}}$, with weights $w_L, w_B$ to enforce physical constraints during optimization.}\label{fig:PINFIG}
\end{figure}

This formulation enables the approximation of complex physical systems without the need for {full mesh evaluation}. Applications of PINNs have been reported in various domains~\cite{Queiroz2021,Pang2019,Raissi2020}, in part due to their ability to handle multiscale behavior and inverse problems. {In particular, because PINNs enforce the governing equations at sampled collocation points rather than on a fixed computational grid, they can alleviate the grid-based manifestation of the curse of dimensionality that affects many traditional numerical methods}~\cite{Cuomo2022}   . Given these properties, PINNs may offer a promising path toward quantum algorithms for solving complex PDEs such as the Navier–Stokes equations. Encoding a PINN into a quantum neural network (QNN) could help mitigate bottlenecks in existing quantum approaches.

Quantum neural networks (QNNs) remain an emerging technology with significant challenges~\cite{Garg2020}, yet recent studies suggest they hold promise for machine learning and quantum artificial intelligence~\cite{Abbas2021}. Here, we explore whether QNNs can represent PINNs in a manner compatible with current and near-term quantum hardware \cite{Berger2025}.

\section{\label{sec:QPINN} ADAPTIVE QUANTUM PHYSICS-INFORMED NEURAL NETWORKS (AQPINN)}

Quantum Neural Networks (QNNs) are the quantum analog of classical neural networks, {and in the near-term formulation considered here, are implemented as parameterized quantum circuits acting on quantum states}. Like their classical counterparts, QNNs aim to approximate functions and recognize patterns from input data by solving an optimization problem. In this context, {the trainable quantum layer is defined through quantum operations applied to quantum states, with gate parameters playing a role analogous to trainable weights in classical networks}. There are various formulations of QNNs in the literature \cite{Tacchino2021,Macaluso2020,Abbas2021,Tacchino2020}, {including quantum-neuron models and parameterized quantum-circuit formulations}\cite{Garcia2022,Macaluso2020}. In {the circuit-based formulation used here}, the input data is encoded into a quantum state $|\mathbf{x}\rangle$ and manipulated through quantum rotation gates, typically denoted $R(\theta)$, and entangling operations such as Controlled-NOT. These operations form the basis for parameterized quantum circuits, which define the overall structure of the QNN. A QNN can be organized in a layered architecture analogous to classical neural networks, where individual quantum neurons are arranged into input, hidden, and output layers. The quantum state flows through the network according to the defined connectivity of its neurons. The training process involves optimizing the rotation angles $\theta$ in the parameterized gates using a global cost function, which is minimized via variational techniques.

The integration of quantum computing with PINNs raises a compelling question: what advantages could Quantum Physics-Informed Neural Networks (QPINNs) offer for solving complex PDEs, such as the Navier–Stokes equations? By embedding the PINN methodology within a quantum neural network, QPINNs have the potential to exploit quantum parallelism and the high-dimensional structure of Hilbert spaces, enhancing the expressivity, scalability, and efficiency of physics-informed learning \cite{Berger2025}. {However, although QPINNs may benefit from the expressive structure of parameterized quantum circuits and the compact representation of quantum states, practical advantages in scalability or efficiency remain problem-dependent and are not yet generally established.}

Recent developments in QNNs have attracted significant attention within the quantum computing community \cite{Garg2020}. While substantial technical challenges remain, the intrinsic properties of QNNs, such as their ability to encode complex functions in superposition states, position them as promising candidates for near-term quantum {machine learning} applications \cite{Abbas2021}. QPINNs represent an emerging computational paradigm that fuses the principles of quantum computing with the framework of PINNs to address PDEs and other complex physical systems \cite{Berger2025}. A possible advantage, though unproven, is the exponential representational capacity of quantum circuits, which could allow high-dimensional functions to be encoded in superposition states. This might enable QPINNs to approximate intricate solutions with significantly fewer parameters than classical networks \cite{Abbas2021}.

\subsection{\label{sec:LA} METHODOLOGY AND ALGORITHM}

We propose a hybrid Adaptive Quantum Physics-Informed Neural Network (AQPINN) framework for solving nonlinear partial differential equations, featuring two key enhancements over conventional PINNs. First, we introduce an adaptive collocation point sampling strategy that builds upon the residual-based probability density approach of \cite{wu2022comprehensive}, but extends it by enabling individual points to move, spawn, or be pruned with memory: points are relocated via gradient-driven dynamics, undergo Gaussian-noise fission near persistent high-residual regions, and are removed through importance-based pruning. Second, we develop an attention-based loss balancing mechanism, inspired by \cite{mcclenny2020self}, which learns global weights—one per loss term (physics, boundary, and initial)—using a simple softmax parameterization trained via ordinary gradient descent, in contrast to pointwise minimax gradient ascent. These innovations directly address the well-known limitations of uniform sampling and static loss weighting in standard PINNs, particularly in regions with sharp gradients and localized solution features. The overall algorithm is grounded in a hybrid quantum–classical neural network architecture, which we detail in the following. In addition, an ablation analysis quantifying the contribution of each adaptive mechanism is provided in Appendix A. A comparison between AQPINN and an equivalent classical adaptive PINN (APINN) is beyond the scope of the present work. Such a comparison would address whether quantum circuits provide additional benefit beyond classical networks when both employ adaptive strategies. While prior studies have shown QPINNs can outperform classical PINNs in specific settings, extending these comparisons to the adaptive setting is an important direction for future research. Here, we focus on the orthogonal question of whether adaptive strategies improve QPINN performance, independent of the classical vs. quantum comparison.

\subsection{HYBRID QUANTUM-CLASSICAL NEURAL NETWORK FORMULATION}

For the AQPINN formulation, we define the solution ansatz \(u_\theta(t,x)\) via a hybrid quantum-classical architecture. {For notational simplicity, we present the AQPINN formulation for a scalar solution
\(u_\theta(t,x)\) that depends on one spatial coordinate and time. The extension to
multidimensional inputs and vector-valued solution fields is straightforward and
is obtained by replacing \((t,x)\) with a general coordinate vector
\(\boldsymbol{\xi}\in\mathbb{R}^{d_{\rm in}}\) and by allowing the final
postprocessing layer to output \(\mathbf{u}_\theta(\boldsymbol{\xi})\in
\mathbb{R}^{d_u}\)}. Given the spatiotemporal input \(\mathbf{s}^{(0)} = (t,x)^\top \in \mathbb{R}^2\), the forward pass consists of a classical preprocessing block, a subsequent quantum circuit, and a final classical postprocessing block. The specific composition and function of each of these three components are described below.

The classical preprocessing network consists of $L_c^{\text{pre}}$ layers:
\begin{equation}
\mathbf{s}^{(\ell+1)} = \sigma\left( W^{(\ell)} \mathbf{s}^{(\ell)} + \mathbf{b}^{(\ell)} \right),
\quad \ell = 0, \dots, L_c^{\text{pre}}-1,
\end{equation}
where $\sigma(\cdot)$ is a nonlinear activation function (e.g., SiLU), $W^{(\ell)} \in \mathbb{R}^{d_{\ell+1} \times d_\ell}$, and $\mathbf{b}^{(\ell)} \in \mathbb{R}^{d_{\ell+1}}$. The output of the preprocessing stage is $\mathbf{z} = \mathbf{s}^{(L_c^{\text{pre}})} \in \mathbb{R}^{n_q}$. 
The quantum layer defines a nonlinear transformation:
\begin{equation}
\mathbf{q} = \mathcal{Q}_\theta(\mathbf{z}) \in \mathbb{R}^{n_q},
\end{equation}
where each component is given by the expectation value:
\begin{equation}
q_i(\mathbf{z};\theta_q) = \langle 0^{\otimes n_q} | U^\dagger(\mathbf{z},\theta_q)\, Z_i \, U(\mathbf{z},\theta_q) | 0^{\otimes n_q} \rangle.
\end{equation}
In the quantum circuit, the unitary operator $U$ is defined as:
\begin{equation}
U(\mathbf{z},\theta_q) = \prod_{\ell=1}^{L_q} U_{\text{ent}}^{(\ell)} \, U_{\text{rot}}^{(\ell)}(\theta_q^{(\ell)}) \, U_{\text{enc}}(\mathbf{z}).
\end{equation}
{The operator consists of three steps}: first, the data encoding,$U_{\text{enc}}(\mathbf{z}) = \prod_{i=1}^{n_q} R_X(z_i) R_Z(z_i)
$, second the variational rotations $U_{\text{rot}}^{(\ell)} = \prod_{i=1}^{n_q} R_Z(\theta^{(\ell)}_{i,1}) R_Y(\theta^{(\ell)}_{i,2}) R_Z(\theta^{(\ell)}_{i,3})$, and the entangling layer $U_{\text{ent}}^{(\ell)} = \prod_{i=1}^{n_q-1} \text{CNOT}_{i,i+1}
$. Finally, we have the classical postprocessing layers where the quantum outputs are passed through a classical network:
\begin{equation}
\mathbf{s}^{(\ell+1)} = \sigma\left( W^{(\ell)} \mathbf{s}^{(\ell)} + \mathbf{b}^{(\ell)} \right),
\quad \ell = L_c^{\text{pre}}, \dots, L_c-1,
\end{equation}
with input $\mathbf{s}^{(L_c^{\text{pre}})} = \mathbf{q}$. The final output is:
\begin{equation}
u_\theta(t,x) = \mathbf{s}^{(L_c)} \in \mathbb{R}.
\end{equation}
In a global composition, the full hybrid model can be written compactly as:
\begin{equation}
u_\theta(t,x) = \mathcal{N}_{\text{post}} \circ \mathcal{Q}_{\theta_q} \circ \mathcal{N}_{\text{pre}} (t,x),
\end{equation}
where $\theta = \{\theta_c, \theta_q\}$ includes classical and quantum parameters. Like a regular classical neural network, the model is trained via gradient-based optimization. Gradients are computed  for the quantum circuit parameters as:
\begin{equation}
\frac{\partial u_\theta}{\partial \theta_q}
=
\frac{\partial u_\theta}{\partial \mathbf{q}}
\cdot
\frac{\partial \mathbf{q}}{\partial \theta_q},
\end{equation}
where $\partial \mathbf{q} / \partial \theta_q$ is evaluated using the parameter-shift rule:
\begin{equation}
\frac{\partial q_i}{\partial \theta}
=
\frac{1}{2}
\left[
q_i(\theta + \frac{\pi}{2}) - q_i(\theta - \frac{\pi}{2})
\right].
\end{equation}

Our goal with this hybrid architecture, interpreted as a nonlinear approximation where the quantum layer acts as a high-dimensional feature map in Hilbert space, is to enhance expressivity beyond classical networks.

\subsection{HYBRID ADAPTIVE QUANTUM-CLASSICAL PINN ALGORITHM}

The AQPINN architecture incorporates a parameterized quantum circuit as a nonlinear feature map within a classical neural network framework. Input coordinates \((t,x)\) are normalized and projected to a latent space via classical preprocessing layers. The latent vector is then encoded into a variational quantum circuit, wherein a sequence of parameterized single-qubit rotations and entangling operations defines a quantum embedding. Expectation values of Pauli observables are extracted and decoded by a classical postprocessing network to approximate the solution \(u_\theta(t,x)\). The training objective is a composite physics-informed loss encompassing the PDE residual, boundary conditions, and initial conditions. For the present illustration, the residual of Burgers' equation is {computed by differentiating the full hybrid ansatz, for example using automatic differentiation in simulation}:
\begin{equation}
r_\theta(t,x) = \partial_t u_\theta + u_\theta \, \partial_x u_\theta - \nu \, \partial_{xx} u_\theta,
\end{equation}
where $\nu$ denotes the viscosity coefficient. The individual loss terms are defined as
\begin{equation}
\mathcal{L}_p = \frac{1}{|\mathcal{X}_c|} \sum_{(t,x)\in \mathcal{X}_c} r_\theta(t,x)^2,
\end{equation}
\begin{equation}
\mathcal{L}_b = \frac{1}{|\mathcal{X}_b|} \sum_{(t,x)\in \mathcal{X}_b} \left|u_\theta(t,x) - u_b(t,x)\right|^2,
\end{equation}
\begin{equation}
\mathcal{L}_i = \frac{1}{|\mathcal{X}_i|} \sum_{(t,x)\in \mathcal{X}_i} \left|u_\theta(t,x) - u_i(t,x)\right|^2.
\end{equation}
where \(\mathcal{X}_b\), \(\mathcal{X}_i\), and \(\mathcal{X}_c\) are the collocation point sets for the boundary conditions, initial conditions, and PDE residual, respectively. Here, \(|\mathcal{X}|\) denotes the cardinality of the corresponding set. The total loss is formulated as a weighted combination,
\begin{equation}
\mathcal{L} = \omega_p \mathcal{L}_p + \omega_b \mathcal{L}_b + \omega_i \mathcal{L}_i,
\end{equation}
where the weights $(\omega_p, \omega_b, \omega_i)$ are dynamically {parametrized} through a soft attention mechanism,
\begin{equation}
(\omega_p, \omega_b, \omega_i) = \mathrm{softmax}(\boldsymbol{\alpha}).
\end{equation}
This adaptive weighting {is intended to mitigate} imbalance between competing objectives during training.

A key feature is the residual-driven adaptive collocation strategy, which iteratively refines the distribution of training points. At prescribed intervals, the collocation set $\mathcal{X}_c$ is updated through three complementary operations:

\begin{enumerate}
    \item \textbf{Gradient-driven relocation:} Points with large residuals are moved along the gradient direction $\nabla_x (r_\theta^2)$, effectively performing {gradient} ascent in the residual landscape to concentrate sampling in regions of high error.
    
    \item \textbf{Local refinement:} New collocation points are generated in the vicinity of regions exhibiting persistently high residuals, improving resolution around steep gradients and shock-like structures {when present}.
    
    \item \textbf{Pruning:} Points with low residual contributions and large age are removed to maintain computational efficiency and avoid redundancy.
\end{enumerate}

To ensure numerical stability and input-output normalization is performed. In particular, the spatial coordinate is explicitly {centered} to have zero mean, preventing bias in the gradient field that could otherwise lead to asymmetric point distributions. The training procedure follows a two-stage optimization strategy. An initial phase using the Adam optimizer enables rapid exploration of the parameter space, followed by refinement using the L-BFGS algorithm to achieve {improved local convergence}. The adaptive sampling mechanism operates throughout both phases, continuously redistributing collocation points in response to the evolving residual landscape. Overall, the proposed framework integrates {quantum-circuit-based} representations, adaptive sampling, and dynamic loss balancing into a unified methodology. This combination {is designed to improve accuracy and sampling efficiency for nonlinear PDEs, particularly in regimes characterized by steep gradients or multiscale behavior}. For an overall view of this numerical procedure, see Algorithm~\ref{alg:qpinn_adaptive}.

\begin{algorithm}[!ht]
\caption{QPINN with Adaptive Collocation and Attention for Burgers' Equation}
\label{alg:qpinn_adaptive}
\begin{algorithmic}[1]

\REQUIRE Initial collocation points $N_0$, maximum points $N_{\max}$, epochs $E$, learning rates $\eta_{\text{Adam}}, \eta_{\text{LBFGS}}$
\ENSURE Trained QPINN model $u_\theta(t,x)$

\STATE Initialize QPINN parameters $\theta$ (classical + quantum layers)
\STATE Initialize attention weights $\alpha = (\alpha_p, \alpha_b, \alpha_i)$
\STATE Generate initial collocation points $\mathcal{X}_c$ via Latin Hypercube Sampling
\STATE Fit data normalizer using sampled $(t,x)$ and initial condition

\FOR{epoch $=1$ to $E$}

    \STATE Sample boundary data $(\mathcal{X}_b, u_b)$
    \STATE Sample initial condition data $(\mathcal{X}_i, u_i)$
    \STATE Obtain trainable collocation points $\mathcal{X}_c$

    \STATE Compute physics residual:
    \[
    r_\theta = \partial_t u_\theta + u_\theta \partial_x u_\theta - \nu \partial_{xx} u_\theta
    \]

    \STATE Compute loss components:
    \[
    \mathcal{L}_p = \frac{1}{|\mathcal{X}_c|} \sum r_\theta^2, \quad
    \mathcal{L}_b = \|u_\theta(\mathcal{X}_b) - u_b\|^2, \quad
    \mathcal{L}_i = \|u_\theta(\mathcal{X}_i) - u_i\|^2
    \]

    \STATE Update attention weights:
    \[
    \omega \leftarrow \text{softmax}(\alpha)
    \]

    \STATE Compute total loss:
    \[
    \mathcal{L} = \omega_p \mathcal{L}_p + \omega_b \mathcal{L}_b + \omega_i \mathcal{L}_i
    \]

    \STATE Update network parameters $\theta$ using optimizer (Adam or L-BFGS)

    \IF{epoch mod $K = 0$}
        \STATE Compute residual gradients $\nabla_x (r_\theta^2)$
        \STATE \textbf{Move} high-residual points toward $\nabla_x (r_\theta^2)$ (specified for each case).
        \STATE \textbf{Add} new points near persistent high-residual regions (specified for each case).
        \STATE \textbf{Remove} aged low-residual points (specified for each case).
        \STATE Enforce domain constraints
    \ENDIF

    \STATE Store loss and attention weights

\ENDFOR

\STATE \textbf{return} trained model $u_\theta(t,x)$

\end{algorithmic}
\end{algorithm}

In our adaptive refinement, the criterion for adjusting the distribution and density of collocation points is driven by a multi-strategy framework based on the spatial magnitude and historical trajectory of the PDE residual. During training, the model periodically computes the physics residual at every collocation point, sorting them while maintaining an exponential moving average (EMA, decay $\alpha = 0.7$) to track historical error trends and smooth temporal oscillations. 

The first strategy actively shifts points via a gradient-ascent mechanism along the positive gradient of the squared residual landscape, $\nabla_{\mathbf{x}} (r^2)$. The algorithm selects high-residual outlier points exceeding a factor of $20.0$ of the global median residual and updates their spatial coordinates according to:
\begin{equation}
\mathbf{x}_i^{(k+1)} = \mathbf{x}_i^{(k)} + \eta \cdot \min\left(r_i, \zeta_{\text{max}}\right) \cdot \frac{\nabla_{\mathbf{x}} (r_i)^2}{\|\nabla_{\mathbf{x}} (r_i)^2\|_2}
\end{equation}\label{adaptEq}
where $\eta$ represents the adaptation learning rate, and the spatial displacement velocity is bounded by a safety cap of $\zeta_{\text{max}} = 0.1$. 
In addition to the advection procedure, a stochastic fission strategy handles localized densification. The model isolates parent points falling within the top $15\%$ of the highest persistent historical errors that have reached a minimum training maturity of 10\% of total epochs. Around each parent point, a candidate pool of 10 structural perturbations is generated by adding random Gaussian noise ($\sigma_{\text{noise}} = 0.05$). To prevent redundant clustering, these new points are appended to the existing collocation set if and only if they maintain a spatial exclusion zone with a minimum Euclidean distance from all pre-existing points, bounding total grid growth to a hard ceiling of $N_{\text{max}}$.

However, in contrast to standard adaptive implementations where points are never removed during training, our advanced scheme introduces an explicit low-importance pruning strategy to manage global memory. Triggered whenever the total point population expands beyond $20\%$ of initial collocation points, active nodes become candidates for removal if their historical residual falls within the bottom $15\%$ of lowest errors and they possess a maturity age exceeding 20\% epochs. Rather than discarding regions indiscriminately that might become problematic later, the algorithm computes a composite importance score (ad-hoc formula):
\begin{equation}
S_i = \bar{r}_i + \frac{0.1}{a_i + 1}
\end{equation}
where $a_i \in \mathbb{N}_0$ represents the point's numerical longevity. This parameter is a discrete survival counter initialized at zero upon point birth, incremented uniformly by one at each adaptation cycle, and fully preserved during spatial advection. This metric mathematically balances a point's localized error against its numerical longevity, and nodes with the lowest $S_i$ scores are systematically removed to free memory slots for emerging high-gradient physical structures.

\section{\label{sec:Res}Results}

To assess the effectiveness of the proposed framework, we consider two representative classes of differential equations frequently encountered in scientific computing: (i) ordinary differential equations (ODEs), encompassing both initial- and boundary-value problems, and (ii) partial differential equations (PDEs) arising in fluid mechanics. Each benchmark is associated with a tailored set of hyperparameters, including network architecture, training duration, and the initial distribution of collocation points. Note that all of the above procedures apply to every test case in this section.

Rather than performing an exhaustive hyperparameter optimization study, our objective is to identify challenging  configuration in which the baseline QPINN exhibits noticeable performance degradation or convergence failure. This critical setup is then adopted as the reference configuration for the Adaptive Quantum Physics-Informed Neural Network (AQPINN), enabling a direct assessment of the improvements introduced by the adaptive refinement strategy. Since the initial collocation set in QPINN defines the maximum spatial resolution available prior to refinement, any subsequent gains achieved by AQPINN can be attributed to its adaptive sampling mechanism rather than to an increase in the baseline discretization density.

The relatively small quantum circuits (3-4 qubits, 1-6 layers) are chosen to balance computational feasibility with demonstrating the effectiveness of our adaptive strategies. Our primary contribution—the adaptive training methodology—is architecture-agnostic and expected to scale to larger circuits. Future work will explore the interplay between circuit expressivity and adaptive training benefits.

\subsection{\label{sec:ODE}Ordinary differential Equation}
In this subsection, we evaluate the proposed framework on two boundary-value problems. For both benchmarks, the models are tested using a minimal hyperparameter configuration, with identical network architectures and training settings employed for both QPINN and AQPINN to ensure a fair and direct comparison.

\subsubsection{\label{sec:IVP}
One-dimensional Helmholtz equation}

To investigate how QPINN and AQPINN address boundary value problems, we consider the one-dimensional Helmholtz equation
\[
u''(x) + u(x) = 0, \quad x \in [-\pi,\pi],
\]
which describes undamped harmonic oscillations, such as acoustic pressure in a closed domain. The analytical solution is \(u_{\text{true}}(x)=\cos(x)\), consistent with the imposed boundary conditions.

The Quantum Physics-Informed Neural Network (QPINN) is a hybrid
classical--quantum model.  Classical pre-processing layers
($\text{Linear} \to \text{SiLU}$, three stages) map $x\in\mathbb{R}$
to an $n_q$-dimensional feature vector ($n_q=3$ qubits).  A variational
quantum circuit encodes each feature via $R_X$/$R_Z$ rotations, applies
parametrised \texttt{Rot} gates and a CNOT entangling ring, and returns
Pauli-$Z$ expectation values.  Three classical post-processing layers
produce the scalar output $\hat{u}(x)$.

The hybrid architecture is implemented in PennyLane using the \texttt{default.qubit} simulator and consists of 3 qubits arranged in a 1-layer variational quantum circuit. The input scalar is encoded through single-qubit \(R_X\) and \(R_Z\) rotations applied to each qubit. Each variational layer comprises parameterized \(\mathrm{Rot}(\theta,\phi,\omega)\) gates followed by a chain of entangling CNOT operations. The circuit output is obtained by measuring the expectation value of the Pauli-\(Z\) operator on each qubit, producing a feature vector of dimension equal to the number of qubits. Importantly, this architecture is kept fixed across all numerical experiments, ensuring that performance differences arise from the training strategy rather than changes in the underlying quantum model. The quantum layer sits inside a classical encoder–decoder: a 2-layer, 1-neuron fully connected preprocessor (Xavier init) feeds the quantum layer, followed by a symmetric postprocessor. Two trainable attention parameters, \(\omega_p\) and \(\omega_b\), are softmax-normalized to dynamically balance physics-informed and boundary/initial losses. The loss informed by physics is evaluated using N=20 collocation points for AQPINN initially; after the AQPINN point updates are complete, N=40. Thus, N=40 collocation points are used for the QPINN. Here, the training is performed using the L-BFGS optimizer with learning rate \(510^{-3}\). The point count is updated at every 10\% interval of the total epochs, adjusting by ±5\% of the total points. Training proceeds in two sequential phases.
In Phase~1 (epochs 0--199), Adam ($\eta=5\times10^{-3}$),
and  Phase~2 (epochs 200--399), the best Adam checkpoint is reloaded and refined with L-BFGS.

\begin{figure}[!ht]
  \centering
  \begin{subfigure}[b]{0.495\textwidth}
    \centering
    \includegraphics[width=\textwidth]{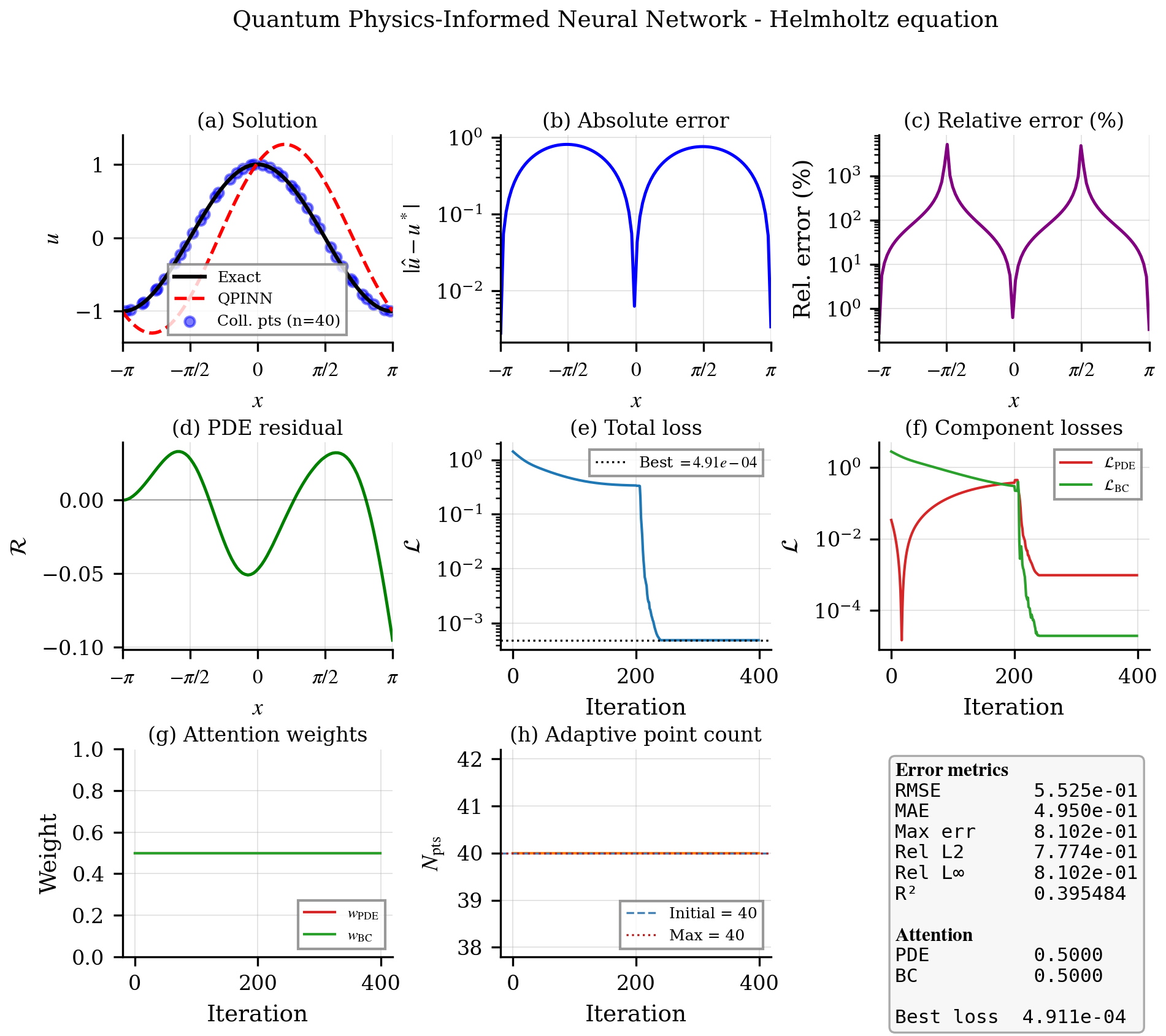}
    \caption{QPINN solution for Helmholtz equation without attention mechanism and adaptive strategy.}
    \label{fig:caseISol}
  \end{subfigure}
  \hfill
   \begin{subfigure}[b]{0.495\textwidth}
    \centering
    \includegraphics[width=\textwidth]{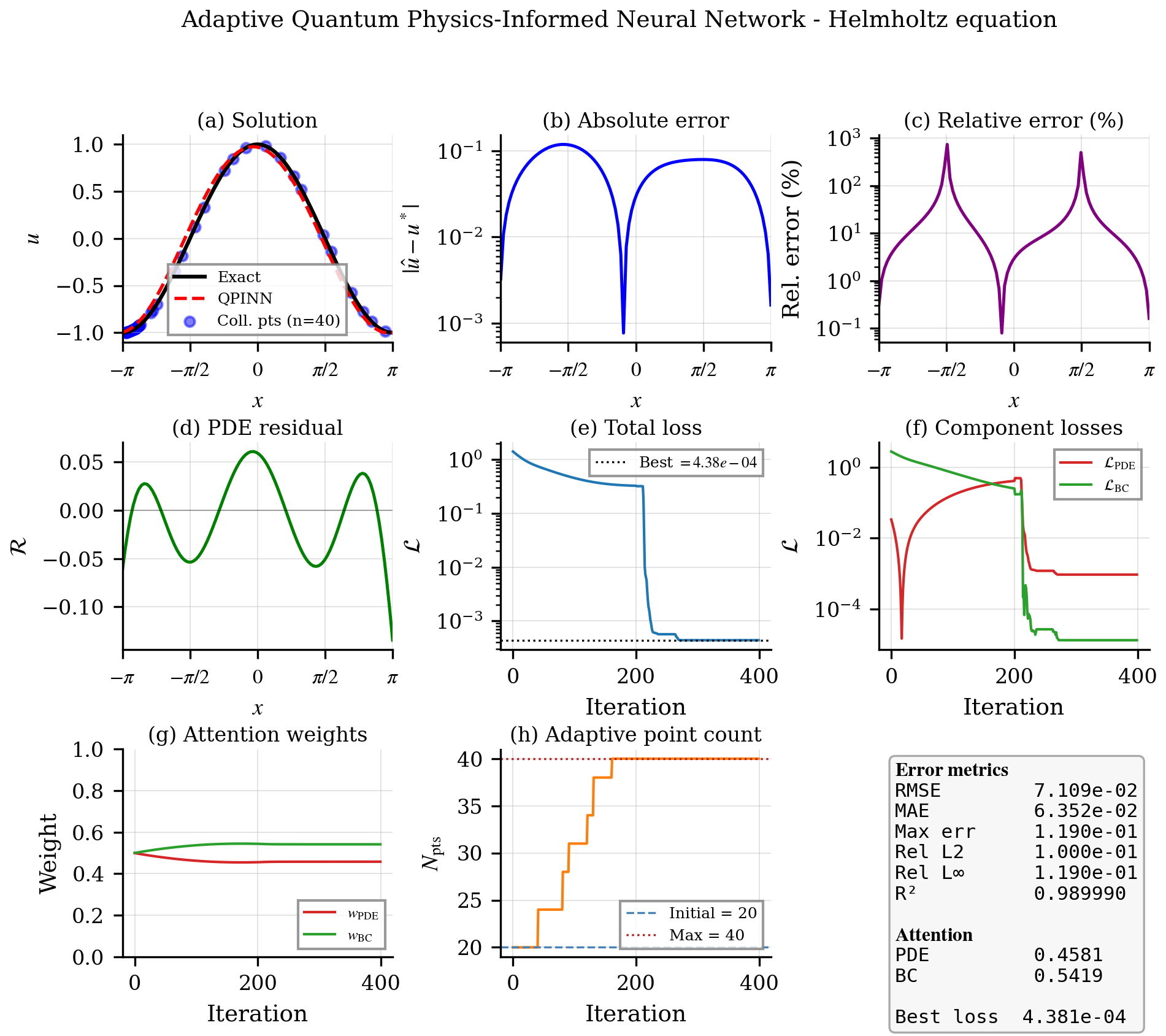}
    \caption{QPINN solution for Helmholtz equation with attention mechanism and adaptive strategy.}
    \label{fig:caseIconv}
  \end{subfigure}
  \caption{Comparison of QPINN and AQPINN solutions for Helmholtz equation.}
  \label{fig:caseI_comparison}
\end{figure}

The comparative performance of the baseline QPINN and AQPINN for solving the Helmholtz equation highlights the critical role of dynamic regularization and self-adaptive sampling in mitigating optimization issues.

Figure~\ref{fig:caseI_comparison} shows that the baseline QPINN fails to accurately reproduce the analytical solution, $u(x)=\cos(x)$ the baseline QPINN fails to capture the correct phase and amplitude of the exact wave solution $u(x)$, yielding an unacceptably low coefficient of determination ($R^2 = 0.395484$) and a maximum absolute error nearing $1.0$. This pathology is characteristic of standard PINN formulations applied to oscillatory PDE operators, where the network becomes trapped in severe local minima. Conversely, the AQPINN framework tracks the exact solution with high fidelity, achieving $R^2 = 0.989990$ and suppressing the absolute error below $10^{-1}$ across the majority of the domain. Notably, both models exhibit localized spikes in relative error exceeding $100\%$ at the roots of the solution where $u(x) \approx 0$, which is an artifact of the vanishing denominator in the relative metric rather than a failure of local convergence.

The underlying optimization dynamics are revealed through the loss trajectories. Both networks exhibit a sharp, discontinuous drop in total loss $\mathcal{L}$ near iteration 200, converging to comparable minima ( $\mathcal{L} \approx 4.91 \times 10^{-4}$ for QPINN and $\mathcal{L} \approx 4.38 \times 10^{-4}$ for AQPINN). However, the baseline QPINN suffers from severe loss imbalance: the boundary condition loss $\mathcal{L}_{b}$ drops rapidly while the PDE residual loss $\mathcal{L}_{p}$ stagnates near $10^{-3}$, forcing the network to satisfy boundary values at the expense of interior physics. AQPINN resolves this multi-objective conflict through the self-adaptive attention mechanism, which dynamically adjusts the weights to $w_{b} = 0.542$ and $w_{p} = 0.458$. This modest rebalancing prevents the boundary gradient from dominating the total backpropagation vector.

Furthermore AQPINN leverages a coarse-to-fine adaptive sampling strategy, initiating training with only $N_{\text{pts}} = 20$ collocation points and progressively allocating points up to the maximum budget of 40 as the residual landscape evolves. This may prevent early-stage overfitting to high-frequency local residuals and allows the network to learn the global wave topology first. The synergy between dynamic loss weighting and adaptive collocation sampling proves essential for navigating the non-convex loss landscapes inherent to quantum and wave-mechanical boundary value problems.
%FALTA INCLUIR FREQUENCIA DE ADAPTABILITADE E A TAXA DE ADAPTABILIDDE COMO É FEITA.

The baseline QPINN uses fixed attention weights, causing the boundary-condition loss to converge rapidly while the physics residual stagnates near $10^{-3}$. In contrast, AQPINN dynamically rebalances the loss terms, improving optimization of the governing equations.

\subsubsection{\label{sec:SMS}
Spring-Mass System problem}

We consider a boundary-value problem requiring a balance between boundary conditions and governing physics, given by a one-dimensional damped, forced harmonic oscillator
\begin{equation}
u'' + 5u' + 6u = 10\sin(x), \quad x \in [0,3],
\label{springmass}
\end{equation}
with initial conditions $u(0)=0$ and $u'(0)=5$. The terms represent inertia, linear damping, and a linear restoring force, respectively, while the forcing term is sinusoidal. The analytical solution is
\begin{equation}
u(x) = -6e^{-3x} + 7e^{-2x} + \sin(x) - \cos(x).
\end{equation}

The QPINN comprises
three sequential stages: a classical preprocessing block, a
parameterized quantum circuit, and a classical postprocessing block.
The scalar input $x$ is first mapped to a feature
vector through two linear layers with one neuron, 
separated by a SiLU activation, with Xavier-normal weight
initialisation and zero biases. These features are then scaled as
encoded qubit-wise into a PennyLane \texttt{default.qubit} circuit of
3 qubits and 3 variational layers. Each layer applies a
general rotation $\mathrm{Rot}(\phi,\theta,\omega)$ to every qubit
followed by a ring of CNOT gates, and the circuit outputs the
expectation values $\langle Z_i\rangle$ ($i=0,\ldots,n_q-1$). A final  two classical layer with one neuron with SiLU maps the quantum outputs to the solution $u(x)$.

The $20$ initial interior collocation points are generated by
Latin Hypercube Sampling (LHS) over $[0,3]$, with a hard cap of
$40$ points. Adaptation is triggered every 10\% of total epochs in both
training phases through three concurrent strategies. Training proceeds in two phases. In Phase~1 (100~Adam epochs), the
main parameters are optimised with Adam ($\eta=10^{-2}$) and
Phase~2 (100~L-BFGS epochs), the best Phase-1 state is restored and
the main parameters are refined with L-BFGS (lr $=0.1$).

\begin{figure}[!ht]
  \centering
  \begin{subfigure}[b]{0.495\textwidth}
    \centering
    \includegraphics[width=\textwidth]{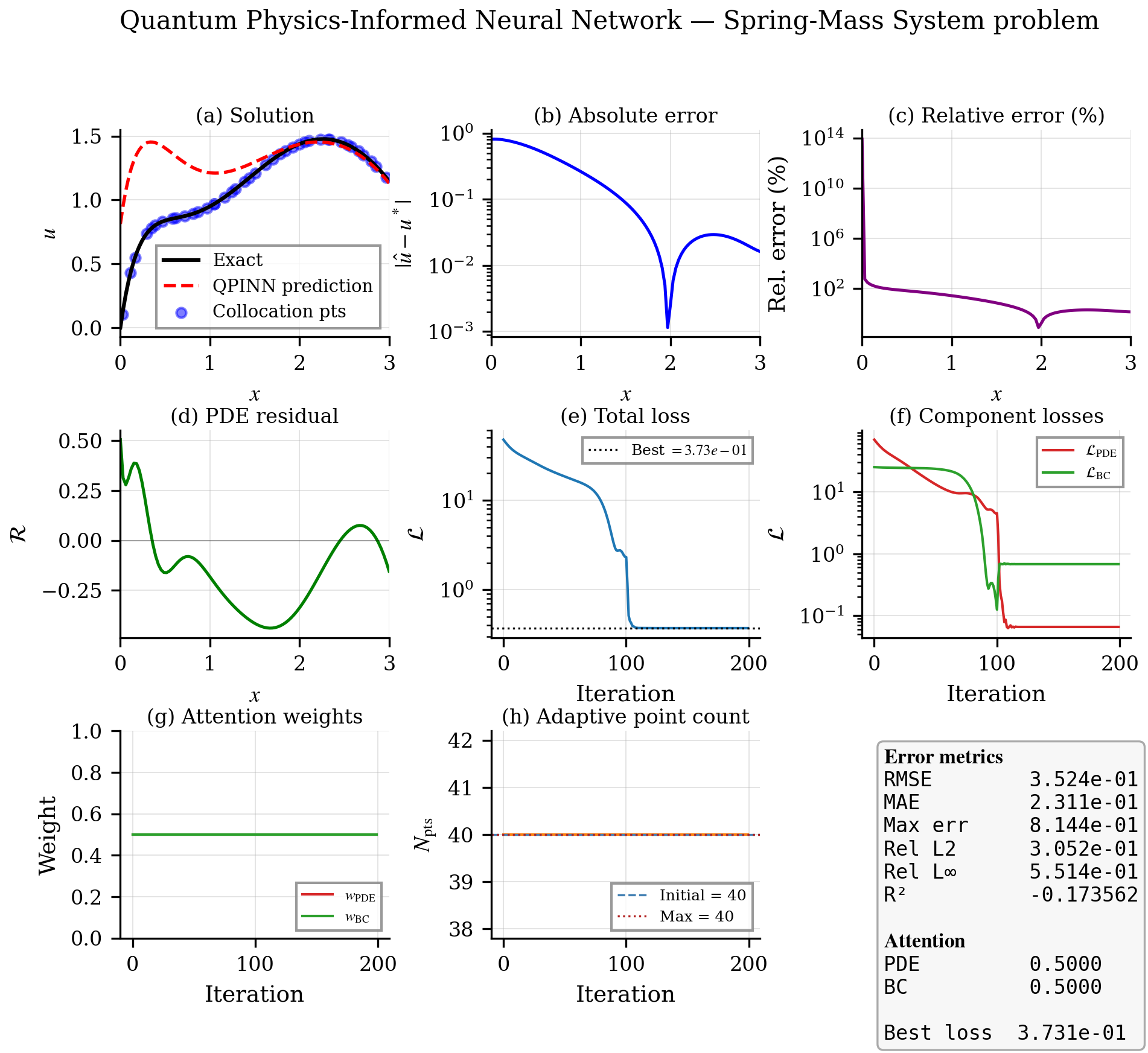}
    \caption{QPINN solution for boundary value problem (spring-mass problem) without attention mechanism and adaptive strategy}
    \label{fig:caseIIconv}
  \end{subfigure}
  \hfill
  \begin{subfigure}[b]{0.495\textwidth}
    \centering
    \includegraphics[width=\textwidth]{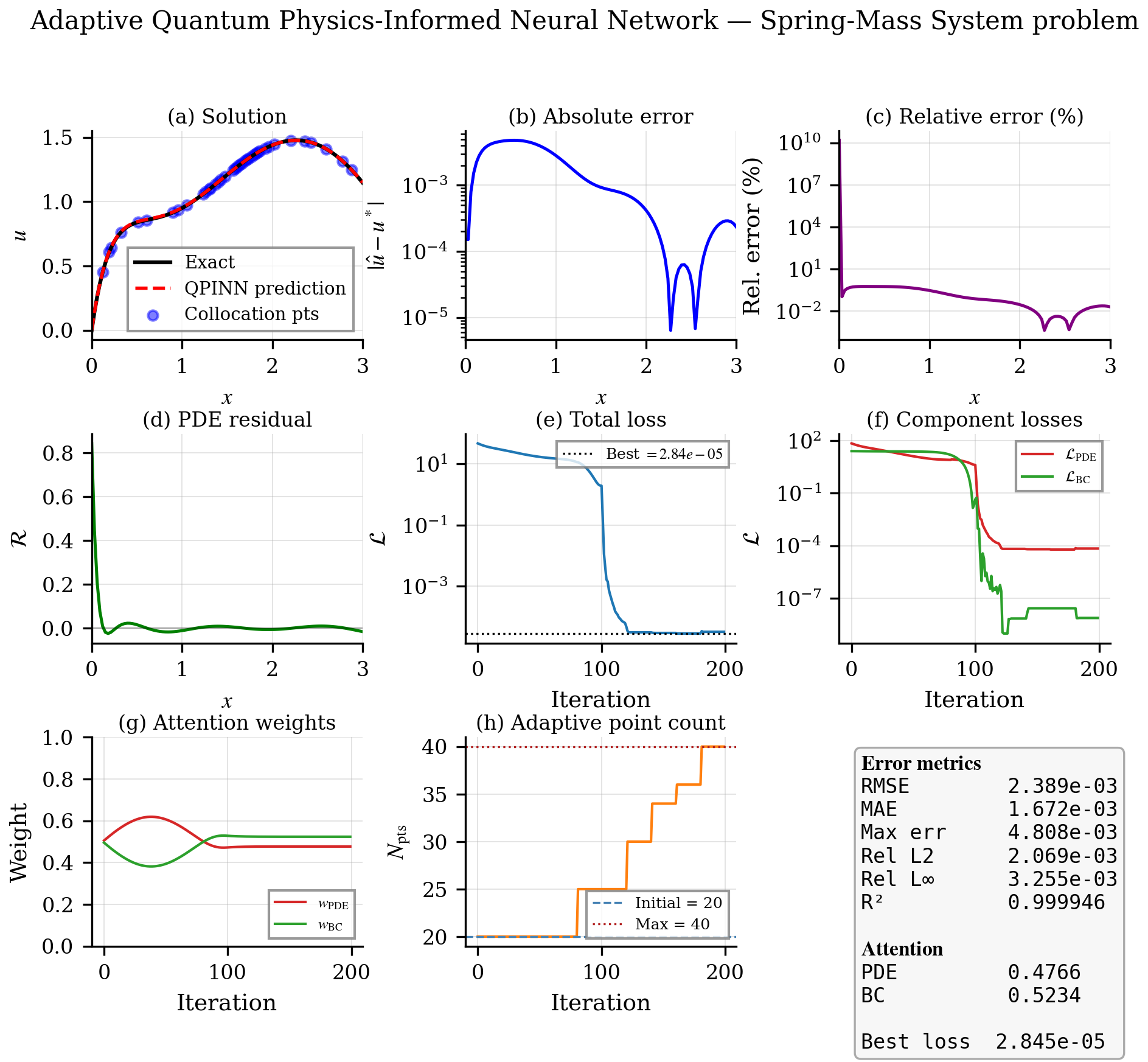}
    \caption{QPINN solution for boundary value problem (spring-mass problem) with attention mechanism and adaptive strategy}
    \label{fig:caseIISol}
  \end{subfigure}
  \caption{Comparison of QPINN and AQPINN solutions for spring-mass problem}
  \label{fig:caseIII_comparison}
\end{figure}

%FALTA INCLUIR PARAMETROS DE ADAPTABILIDADE

To assess the effectiveness of the proposed Adaptive Quantum Physics-Informed Neural Network (AQPINN), we consider a classical spring--mass system and compare its performance against a standard QPINN employing fixed loss weights and a static collocation set. Figure~\ref{fig:caseIII_comparison} summarizes the training dynamics, error evolution, and final solution quality for both approaches.

The baseline QPINN exhibits substantial difficulties in approximating the underlying dynamics, particularly in the region $x\in[0,1]$, where the predicted solution deviates significantly from the analytical reference. This discrepancy is accompanied by large oscillatory residuals, indicating that the governing differential equation is not consistently satisfied throughout the computational domain. Despite continued optimization, the model rapidly reaches a plateau characterized by a relatively large loss value, suggesting the training process becomes trapped in a suboptimal region of parameter space. The fixed weighting strategy, with equal emphasis assigned to the PDE and boundary-condition losses throughout training, limits the model's ability to dynamically focus on the most challenging constraints. Furthermore, the static collocation distribution prevents additional computational effort from being allocated to regions with persistent residual errors. As a consequence, the baseline model attains a final RMSE of $3.5\times10^{-1}$ and a negative coefficient of determination, indicating poor predictive performance and inadequate representation of the physical solution.

In contrast, the adaptive framework produces a solution that closely matches the analytical reference over the entire domain. The error distribution remains uniformly small, with absolute errors typically below $10^{-3}$, demonstrating the effectiveness of the proposed adaptive mechanisms. This improvement arises from the combined action of adaptive loss balancing and residual-based collocation refinement. During training, the attention mechanism continuously adjusts the relative contributions of the PDE and boundary-condition objectives, enabling the optimizer to concentrate on the dominant sources of error at each stage of the learning process. Simultaneously, the adaptive refinement strategy progressively increases the number of collocation points in response to the evolving residual distribution, thereby directing computational resources toward regions requiring higher resolution. The coordinated interaction of these two mechanisms results in a rapid reduction of both component losses and the total objective function, ultimately yielding a best loss of $2.8\times10^{-5}$. The quantitative comparison reported highlights the gains obtained through adaptivity. Relative to the baseline QPINN, the AQPINN reduces the RMSE from $3.5\times10^{-1}$ to $2.3\times10^{-3}$, while the mean absolute error decreases by more than one order of magnitude. Similar improvements are observed for the maximum error and both relative $L_2$ and $L_\infty$ norms. Most notably, the coefficient of determination increases from a negative value to $0.99$, indicating an almost total reconstruction of the analytical solution. 

Likewise, the primary limitation of the standard QPINN is not necessarily the expressive power of the quantum model itself, but rather the inability of a static training strategy to effectively navigate the highly non-convex optimization landscape associated with physics-informed learning. By dynamically reallocating attention between competing loss terms and adaptively enriching the collocation set in regions of elevated residual error, the AQPINN improves convergence stability, solution accuracy, and physical consistency. The observed performance gains suggest that adaptive training mechanisms are essential for scaling quantum physics-informed neural networks to more challenging nonlinear and high-dimensional scientific computing applications.

\subsection{\label{sec:EDP}Partial Differential Equations: Fluid dynamics equations}

In this section, we evaluate representative PDEs arising in Navier--Stokes simulations. The goal is to extend the proposed approach to problems with direct relevance to real-world applications, while using well-established benchmark cases as a controlled setting for numerical assessment and validation. 

\subsubsection{\label{sec:PE}
2D Poisson: Pressure equation}

In this case, we consider the steady-state two-dimensional Poisson equation, which models the pressure field in Navier--Stokes systems, defined at the bounded domain $\Omega = [0,1]^2 \subset \mathbb{R}^2$:
\begin{equation}
\nabla^2 u(x, y) = \frac{\partial^2 u}{\partial x^2} + \frac{\partial^2 u}{\partial y^2} = 0, \quad (x, y) \in \Omega
\end{equation}
subject to non-homogeneous Dirichlet boundary conditions:
\begin{equation}
u(x, 0) = \sin(\pi x), \quad u(x, 1) = 0, \quad u(0, y) = 0, \quad u(1, y) = 0
\end{equation}
The closed-form analytical solution used for verification and error analysis is given by:
\begin{equation}
u_{\text{true}}(x, y) = \frac{\sin(\pi x) \sinh(\pi (1 - y))}{\sinh(\pi)}
\end{equation}

The operational parameters of the framework are structured according to their functional sub-modules. The adaptive collocation scheme initializes with 40 interior points generated via Latin Hypercube Sampling within the domain $\Omega=[0,1]^2$, and enforces a maximum limit of $100$ points to control adaptive growth. During refinement, collocation points are updated using a gradient-based displacement rule based onv$(r_\theta^2)$, with learning rates $\eta \in \{0.02, 0.01\}$. 
The hybrid quantum-classical neural network takes a two-dimensional input $\mathbf{x}=[x,y]^T$  and encodes it through a classical preprocessing and posprocessing network with 2 Layer and 4 neurons with SiLU activations. The encoded features are then passed to a variational quantum circuit with 4 qubits and 6 layers, where parameterized rotations and entangling gates generate expectation values $\langle Z_i \rangle$ as latent representations. The model is trained using adaptive attention weights $[0.4,0.6]$, which balance physics-informed and data-driven loss contributions.
The training setup uses a maximum of 5000 epochs with a 1000-epoch Adam pretraining phase and initial learning rate $10^{-2}$. Boundary conditions are enforced using 160 collocation points (40 per boundary). 

The baseline QPINN model exhibits, according to Fig~\ref{fig:caseIIIconvwithout}, a significant accumulation of spatial errors, with local absolute errors extending up to $3.4 \times 10^{-2}$. These errors concentrate near the boundaries as $x \to 0$ and $x \to 1$, as well as in the upper-middle region of the domain, indicating the failure to simultaneously enforce boundary conditions and interior physics. This behavior is further confirmed by the 1D slice in $y=0.5$, where the predicted solution $u_\theta(x)$ systematically underestimates the peak and deviates near the boundaries. In contrast, the AQPINN model, according to Fig~\ref{fig:caseIIIconvwith}, substantially improves global consistency by flattening the error distribution, reducing the maximum absolute error to $1.4 \times 10^{-2}$. The corresponding 1D profile closely matches the analytical solution across the entire domain. This improvement is reflected in the quantitative metrics, with the RMSE decreasing from $8.2 \times 10^{-3}$ to $2.9 \times 10^{-3}$, the relative error $L^2$ reduced to $1.054$ \% and the determination coefficient reaching $R^2=0.999$.

The performance limitation of the baseline model is primarily explained by the decoupled loss dynamics. Under a uniform weighting scheme with $\omega_{\text{p}}=\omega_{\text{b}}=0.5$, the baseline network stagnates around iteration 1600, with the PDE residual plateauing at approximately $5.0\times10^{-4}$. This indicates convergence to a suboptimal local minimum where the model is unable to further resolve spatial gradients governed by the differential operator. In contrast, the AQPINN formulation introduces a self-adaptive attention mechanism that rapidly rebalances the loss contributions within the first 200 iterations, stabilizing at $\omega_{\text{BC}}\approx0.60$ and $\omega_{\text{PDE}}\approx0.40$. This adaptive shift prioritizes boundary enforcement, mitigating stiffness effects near the domain edges. As a result, the boundary loss is reduced to $3.0\times10^{-5}$, while the PDE residual is simultaneously driven down to $1.0\times10^{-4}$, leading to a more stable and physically consistent optimization trajectory.

The spatial distribution of training nodes provides further insight into the robustness of the AQPINN. The baseline framework uses a static, randomized allocation of 100 points across the domain (see Fig~\ref{fig:caseIIIconvwithout}). This method is fundamentally blind to local residual histories, resulting in an unweighted scatter across regions with highly varying error gradients.

In contrast, the AQPINN utilizes an intelligent spatial re-meshing strategy based on residual history tracking (see Fig~\ref{fig:caseIIIconvwith}). The model automatically detects that the highest residual concentrations occur along the high-gradient boundary regions ($y \to 0$ near $x=0.5$) and within the lateral boundaries ($x \to 0$ and $x \to 1$). It dynamically pulls points away from the low-error upper domain and re-clusters them directly onto these critical regions. This targeted geometric refinement allows the network to resolve local mathematical stiffness without expanding the overall parameter size or requiring an excessively large training dataset. The spatial distribution of the training nodes highlights the advantage of AQPINN over the baseline. The baseline model uses a static random sampling of 100 points, which ignores residual information and leads to inefficient coverage of regions with high error gradients. In contrast, AQPINN performs residual-driven adaptive re-meshing with $100$ points, concentrating points in high-error regions such as boundary layers ($y \to 0$, $x \to 0$, and $x \to 1$) while reducing sampling in low-error zones. This targeted redistribution improves the resolution of local stiffness without increasing the model size or the complexity of the data set.

\begin{figure}[!ht]  % h = here, t = top, b = bottom, p = page of floats
  \centering
  \includegraphics[width=0.98\textwidth]{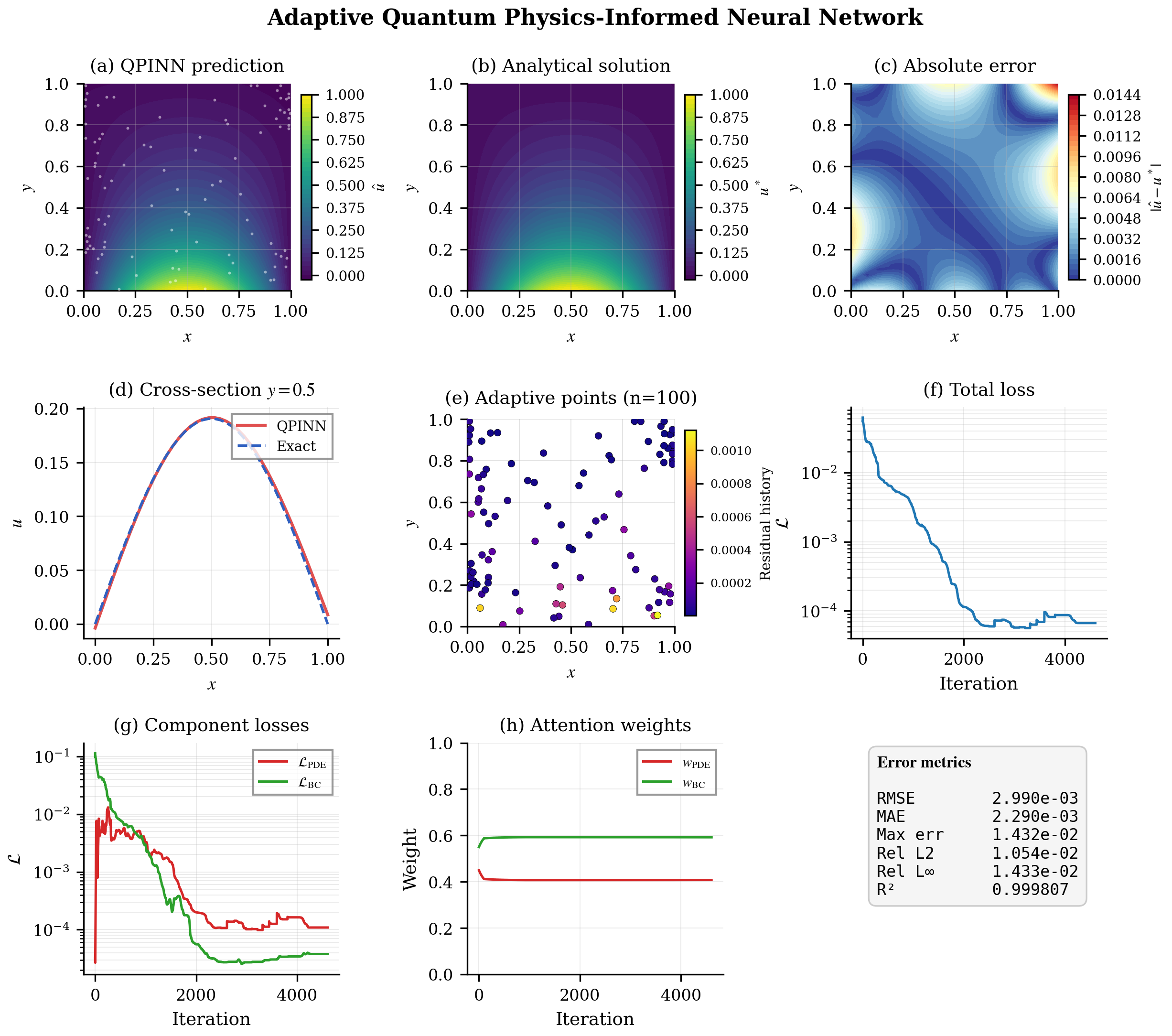}
    \caption{AQPINN solution for Poisson equation  with attention mechanism and adaptive strategy}
    \label{fig:caseIIIconvwith}
\end{figure}

\begin{figure}[!ht]  % h = here, t = top, b = bottom, p = page of floats
  \centering
  \includegraphics[width=0.98\textwidth]{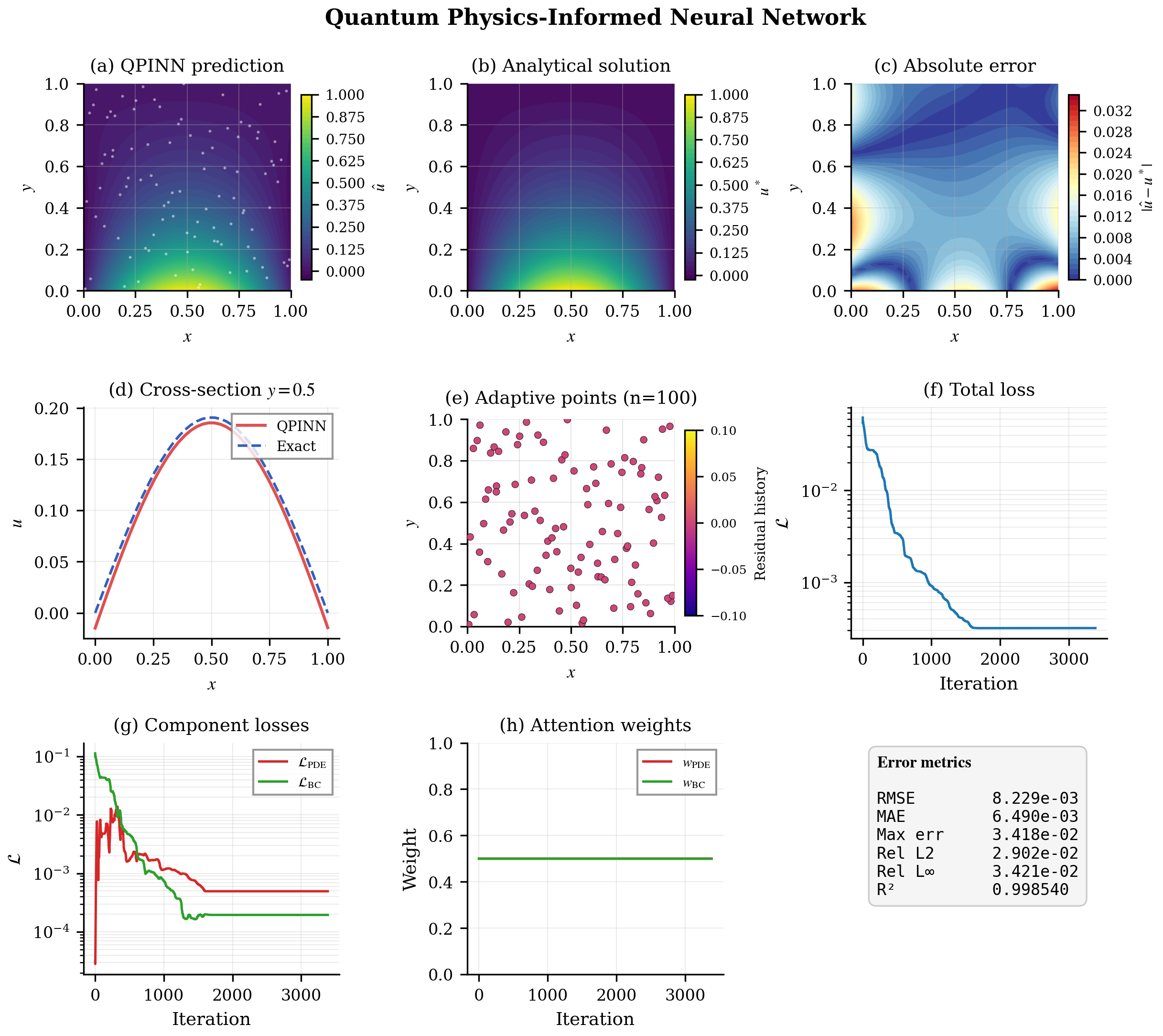}
    \caption{QPINN solution for Poisson equation.}
    \label{fig:caseIIIconvwithout}
\end{figure}

\begin{figure}[!ht]  % h = here, t = top, b = bottom, p = page of floats
  \centering
  \includegraphics[width=0.98\textwidth]{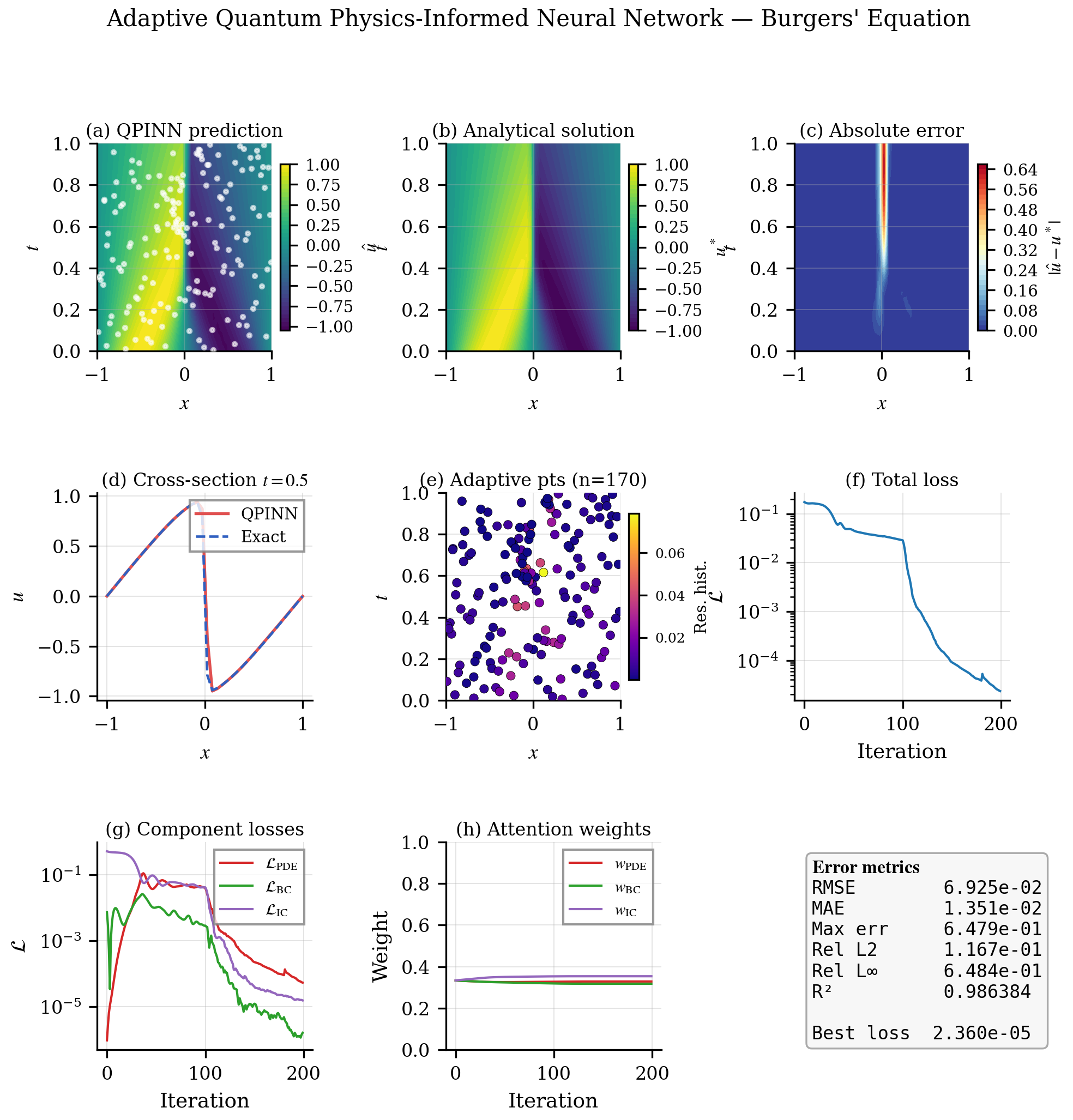}
    \caption{QPINN solution for Burgers' Equation attention mechanism and adaptive strategy.}
    \label{fig:caseIVconvwith}
\end{figure}

\begin{figure}[!ht]  % h = here, t = top, b = bottom, p = page of floats
  \centering
  \includegraphics[width=0.98\textwidth]{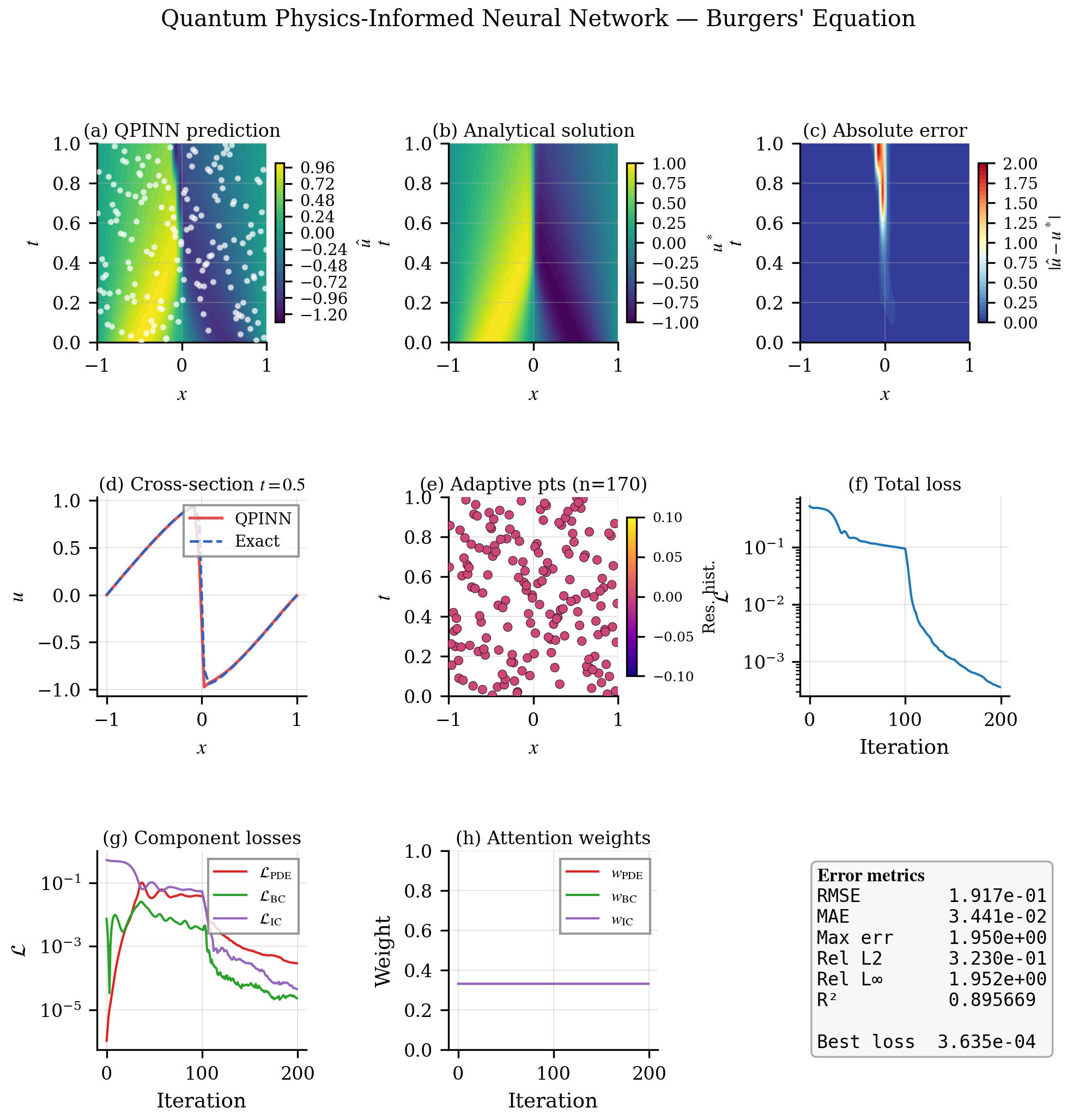}
    \caption{QPINN solution for Burgers' Equation without attention mechanism and adaptive strategy}
    \label{fig:caseIVconvwithout}
\end{figure}

\begin{figure}[!ht]  % h = here, t = top, b = bottom, p = page of floats
  \centering
  \includegraphics[width=0.98\textwidth]{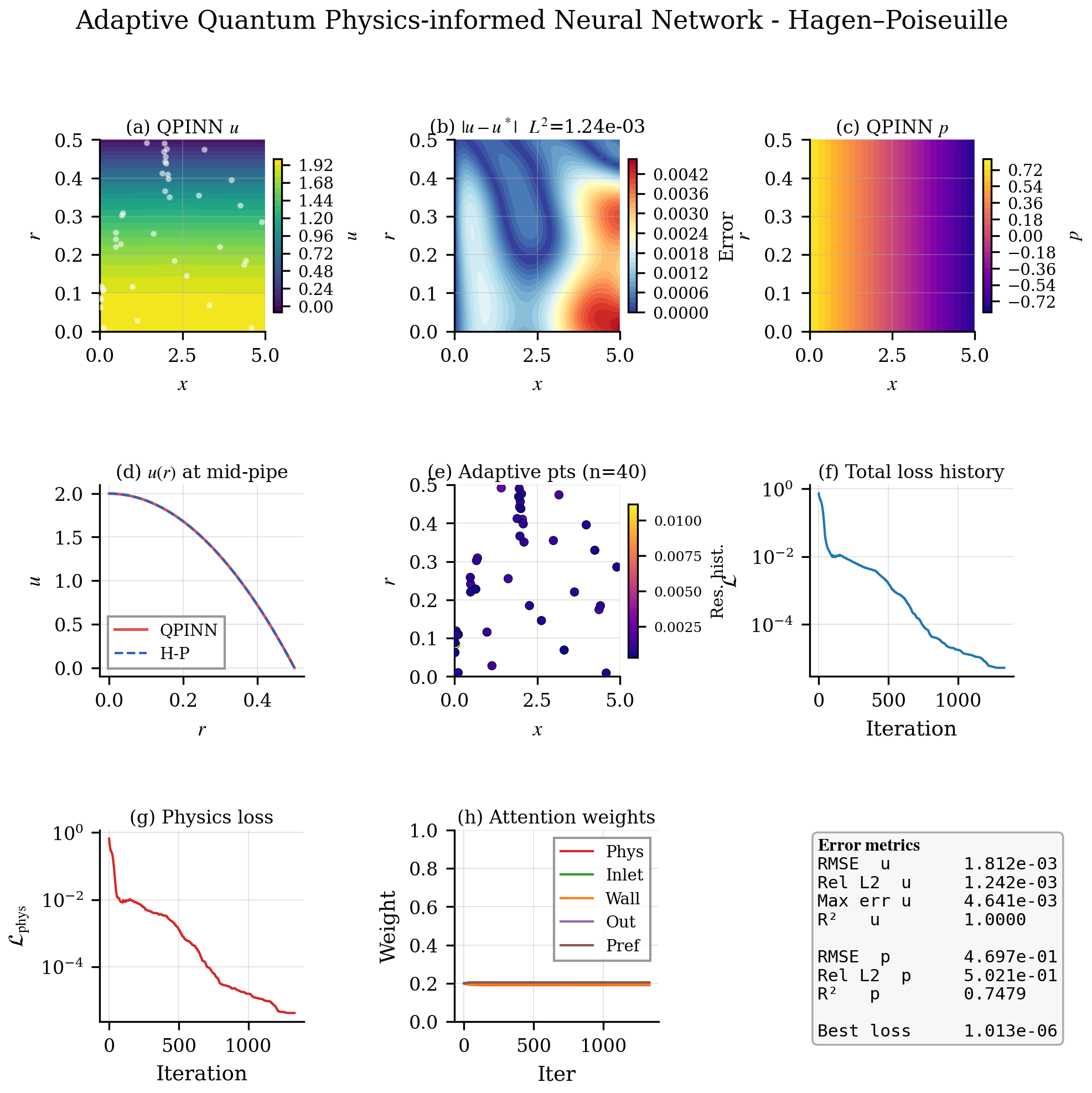}
    \caption{QPINN solution for Hagen-Poiseuille flow with attention mechanism and adaptive strategy}
    \label{fig:caseVconvwith}
\end{figure}

\begin{figure}[!ht]  % h = here, t = top, b = bottom, p = page of floats
  \centering
  \includegraphics[width=0.98\textwidth]{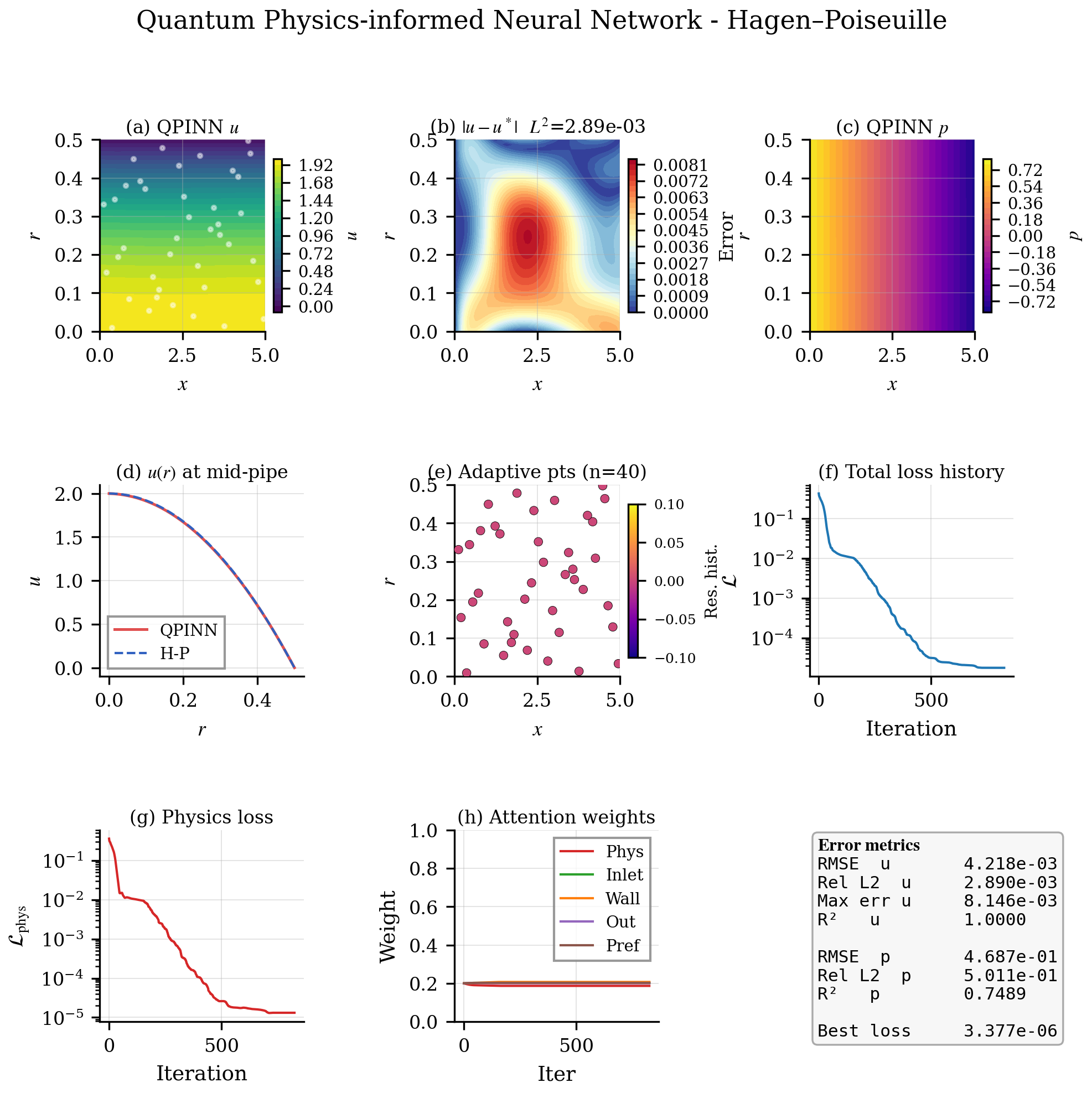}
    \caption{QPINN solution for Hagen-Poiseuille flow }
    \label{fig:caseVconvwithout}
    \end{figure}

\begin{figure}[!ht]  % h = here, t = top, b = bottom, p = page of floats
  \centering
  \includegraphics[width=0.98\textwidth]{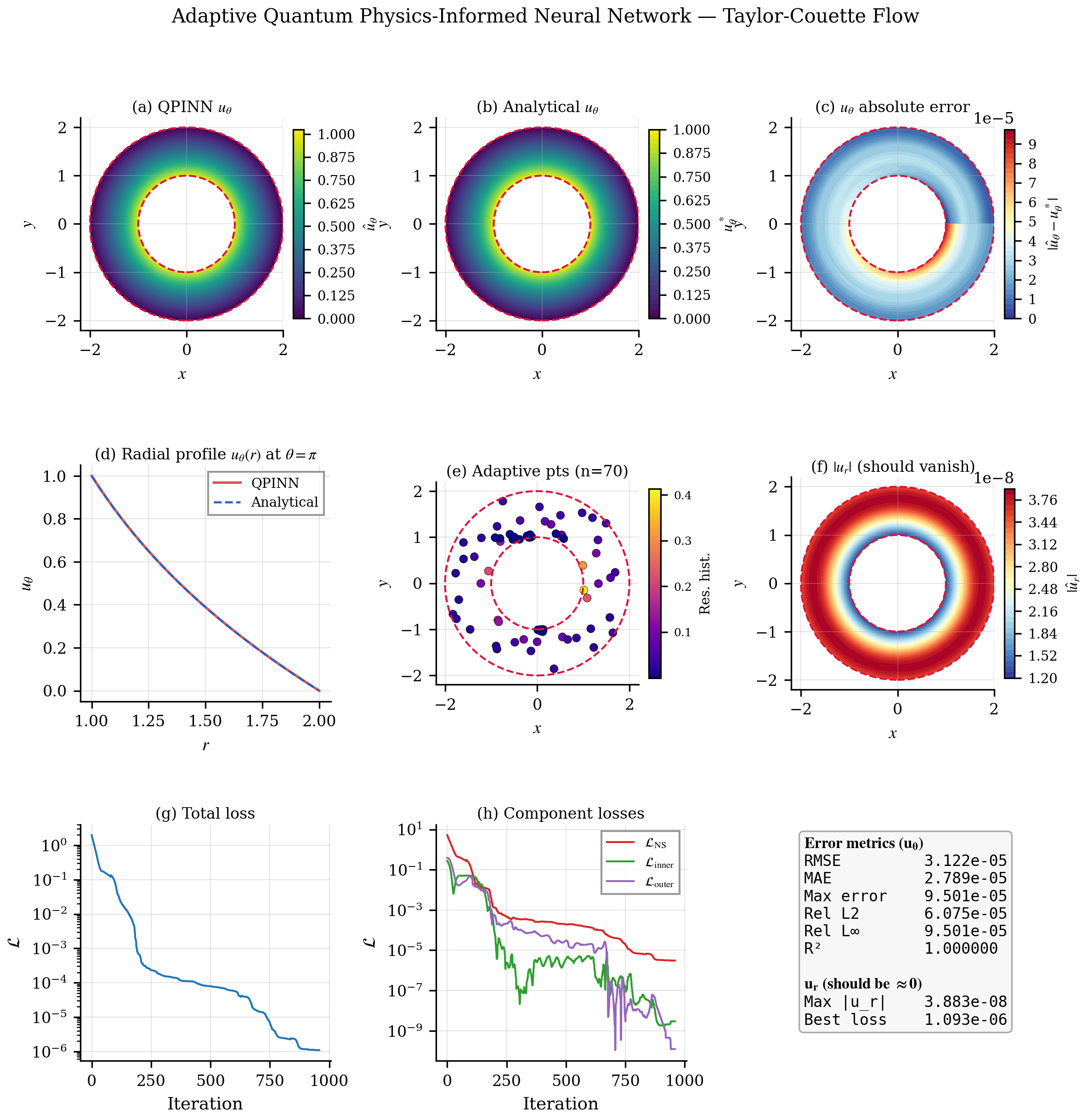}
    \caption{QPINN solution for Taylor-Couette flow with attention mechanism and adaptive strategy}
    \label{fig:casevIconv}
\end{figure}

\begin{figure}[!ht]  % h = here, t = top, b = bottom, p = page of floats
  \centering
  \includegraphics[width=0.98\textwidth]{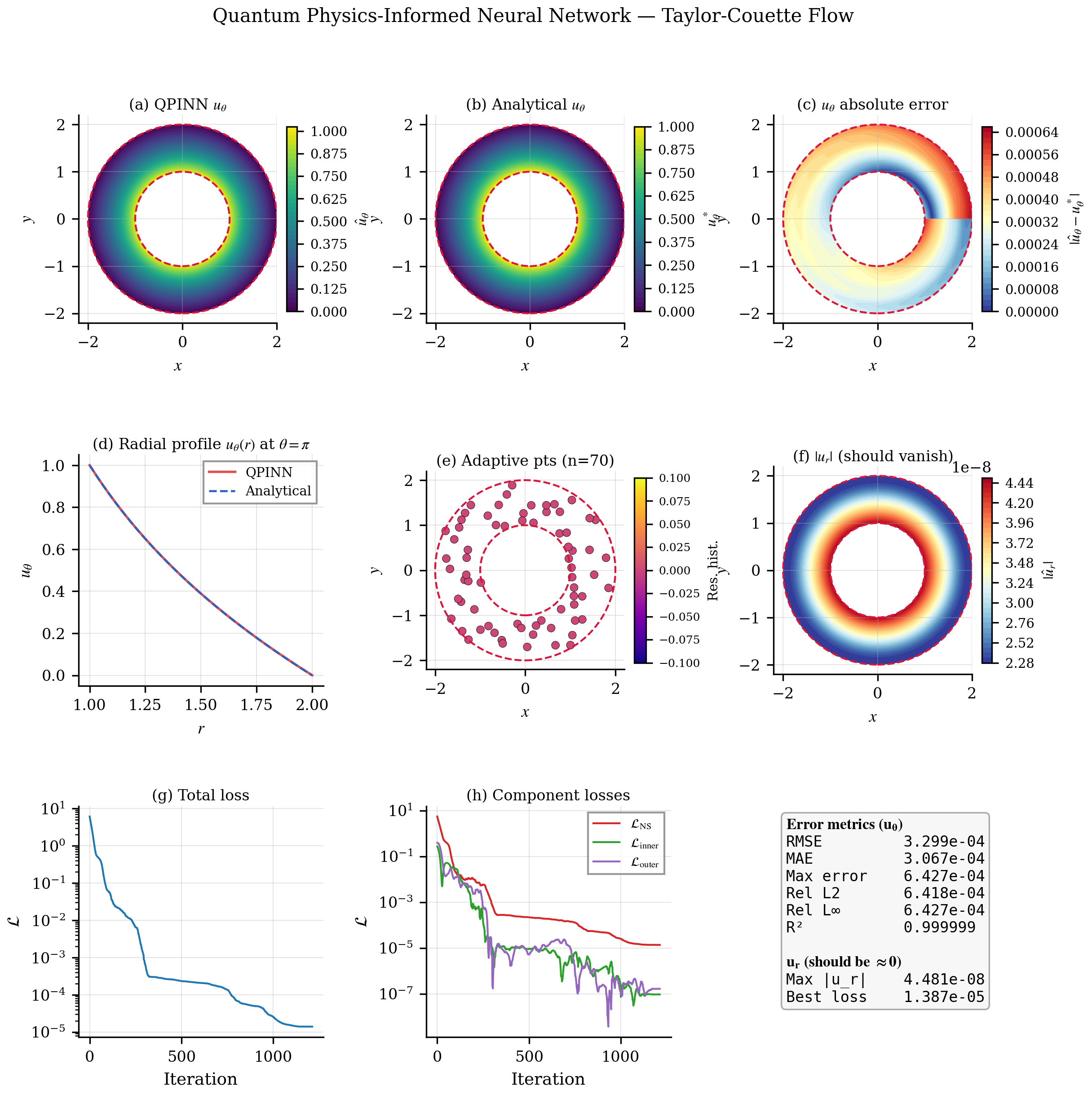}
    \caption{QPINN solution for Taylor-Couette flow.}
    \label{fig:caseVIconv}
\end{figure}

\subsubsection{\label{sec:BE} Burgers' Equation}

To analyze shock effects on QPINN solution quality, we solve the viscous one-dimensional Burgers' equation on the bounded space-time domain $\Omega = [0,1]\times[-1,1]$:
\begin{equation}
\frac{\partial u}{\partial t} + u \frac{\partial u}{\partial x} - \nu \frac{\partial^2 u}{\partial x^2} = 0,
\quad (t,x)\in\Omega,
\end{equation}
with viscosity $\nu = 0.01$. The system is subject to the initial condition $u(0,x) = -\sin(\pi x)$ and homogeneous Dirichlet boundary conditions $u(t,\pm1)=0$. Model accuracy is evaluated against a semi-analytical reference solution obtained via the Cole--Hopf transformation and numerical quadrature.
The network architecture consists of a classical preprocessing MLP with structure 2 Layers of 16 neurons and SiLU activations, followed by a quantum circuit comprising state preparation into parameterized rotation gate with 4 layers and 4 qubits, ending with a classical decoding MLP with 2 Layers of 16 neurons that reconstructs the velocity field. 
The adaptive strategy is again employed and redistributes collocation points according to $\nabla_{\mathbf{x}} (r_i)^2$, removing low-residual samples and generating new points in regions of persistent high error tracked via an exponential moving average.

The residual based adaptive collocation scheme is defined over a space-time domain $[0,1]\times[-1,1]$, where an initial set of $150$ collocation nodes is generated via Latin Hypercube Sampling, subject to a hard upper bound of $200$ nodes to control refinement complexity. A minimum inter-point distance of $0.01$ is enforced to avoid clustering and ensure spatial diversity. Node evolution follows a gradient-driven displacement rule, with learning rates $\eta=0.02$ for exploration and $0.01$ for stabilization phases. Local refinement is further regularized through stochastic perturbations with Gaussian noise of scale $\sigma=0.05$, promoting candidate diversity in high-residual regions. Temporal consistency of the refinement process is maintained using an exponential moving average of the squared residual, ensuring stable tracking of persistent error structures while suppressing transient fluctuations.

The spatial prediction fields indicate that both models attempt to reconstruct the discontinuous structure characteristic of convective–diffusive shock fronts (See Figures~    \ref{fig:caseIVconvwith} and     \ref{fig:caseIVconvwithout}
). However, the absolute error field $|\hat{u}-u^*|$ reveals a pronounced limitation of the baseline QPINN, which exhibits strong error accumulation along the internal discontinuity near $x \approx 0$ for $t>0.4$, reaching a maximum deviation of $1.95$. This behavior is associated with shock smearing, where insufficient local representational capacity leads to over-smoothing of the sharp transition, resulting in a displaced and diffused wave front. The deficiency is further reflected in a reduced predictive quality, with $R^2=0.9$ and a relative $L^2$ error of $32.30\%$. In contrast, the AQPINN effectively suppresses error propagation along the moving front, reducing the maximum absolute error to $0.65$ (a $66.8\%$ decrease) and accurately capturing the sharp transition at $x=0$ at $t=0.5$. This improvement is also consistent globally, where the RMSE decreases from $1.9\times10^{-1}$ to $6.9\times10^{-2}$ and the coefficient of determination increases to $R^2=0.98$, indicating a substantially more stable and physically consistent reconstruction of the shock dynamics.

The optimization behavior can be understood by examining the decoupled loss components. For Burgers’ equation, the resulting landscape is highly non-convex, with strong competition between physics, initial, and boundary constraints. In the baseline QPINN, fixed uniform weights ($w_{p} = w_{b} = w_{i} \approx 0.333$) lead to a stagnation around iteration $\sim 100$, where $\mathcal{L}_{p}$ stalls at $\mathcal{O}(10^{-3})$, reflecting the inability of the optimizer to escape a shock-induced barrier.

In contrast, AQPINN introduces dynamic loss rebalancing that detects gradient conflicts and reallocates weights in favor of $\mathcal{L}_{p}$ and $\mathcal{L}_{i}$ over $\mathcal{L}_{b}$, enabling continued descent beyond this plateau. This adaptive strategy reduces $\mathcal{L}{p}$ to $\mathcal{O}(10^{-4})$ and $\mathcal{L}_{i/b}$ to $\mathcal{O}(10^{-6})$, effectively overcoming the optimization barrier.

Additionally, AQPINN employs residual-driven adaptive collocation, concentrating points along the shock region ($x \approx 0$), unlike the uniform sampling of the baseline that under-resolves high-gradient areas. This targeted re-meshing improves resolution of the discontinuity without increasing the total number of collocation points, highlighting the importance of adaptive sampling in nonlinear, convection-dominated regimes.

\subsubsection{\label{sec:NS-HP}
Navier-Stokes equations - Hagen-Poiseuille flow}

We consider the steady, axisymmetric incompressible Navier–Stokes equations for fully developed flow in a cylindrical pipe, where the velocity field reduces to $u = u(r)$ and $v=0$. The governing momentum equation simplifies to
\begin{equation}
-\frac{1}{\rho}\frac{dP}{dx} + \nu \left( \frac{\partial^2 u}{\partial r^2} + \frac{1}{r}\frac{\partial u}{\partial r} \right) = 0.
\end{equation}

Using characteristic scales $R$ and $U_0$ for length and velocity, respectively, the non-dimensional form over $\Omega = \{(x,r)\mid x \in [0,L],\, r \in [0,R]\} \subset \mathbb{R}^2$ becomes
\begin{equation}
-\frac{\partial p}{\partial x} + \frac{1}{Re}\left( \frac{\partial^2 u}{\partial r^2} + \frac{1}{r}\frac{\partial u}{\partial r} \right) = 0,
\end{equation}
where $Re = \frac{U_0 R}{\nu}$ is $10$.

The system is subject to the boundary conditions
\begin{align}
u(0,r) &= 1, \\
u(x,R) &= 0, \\
\left.\frac{\partial u}{\partial r}\right|_{r=0} &= 0.
\end{align}

The QNN maps normalized spatial coordinates $(x,r)$ through a classical encoder MLP with 2 Layers and 16 neurons, and SiLU activations into a latent quantum feature space. This is processed by a Quantum Circuit with $4$ layers of parameterized rotations, and periodic CNOT entanglement with 4 qubits, followed by a classical decoder MLP with 2 Layer and 16 neurons, and SiLU activations that reconstructs the velocity $u(x,r)$ from Pauli-$Z$ expectations.

Training stability is enhanced via a self-adaptive attention mechanism that learns loss weights scheme, combined with a residual-based adaptive collocation strategy. 
Our residual based adaptive collocation strategy is initialized with 20 interior points generated via Latin Hypercube Sampling, with a maximum cap of 40 points to control dynamic node growth. A minimum Euclidean separation of 0.05 is enforced between collocation points to avoid clustering. The computational domain is defined as $[0,5.0]\times[0,1.0]$, representing the pipe geometry in axial and radial coordinates. Node refinement is driven by a gradient-ascent update rule with learning rates of 0.01 and 0.005. Stochastic exploration is introduced via Gaussian perturbations with standard deviation $\sigma=0.05$ to enhance spatial coverage. Residual tracking is stabilized using an exponential moving average, ensuring temporal againg consistency in the adaptive sampling process.

The global optimization procedure is performed over 300 training epochs using the Adam optimizer, with an initial learning rate of $5\times10^{-3}$ applied jointly to the network parameters. 

An analysis of the reconstructed flow fields reveals a clear divergence in multi-field predictive fidelity between the two frameworks. Both models accurately recover the parabolic velocity profile $u$ at the mid-pipe cross-section ($x = 2.5$), in agreement with the analytical Hagen--Poiseuille solution (see Figures~\ref{fig:caseVconvwith} and     \ref{fig:caseVconvwithout}). The baseline QPINN attains a velocity RMSE of $4.2 \times 10^{-3}$, whereas the AQPINN reduces this error to $1.8 \times 10^{-3}$, with both approaches achieving high coefficients of determination. However, a severe degradation is observed in the baseline pressure field prediction, which exhibits a non-physical distribution (see Figures~\ref{fig:caseVconvwithout}), characterized by a relative $L^2$ error of $50.11\%$ and a reduced fit quality ($R_p^2 = 0.75$). This failure propagates into the velocity solution, manifesting as a pronounced error pocket in the domain core ($x \approx 2.5, r \approx 0.25$) with a peak absolute deviation of $8.2 \times 10^{-3}$, indicating that inaccurate pressure gradients strongly bias the velocity optimization away from physical consistency. In contrast, the AQPINN, according to Fig~    \ref{fig:caseVconvwith}, effectively mitigates this coupled error propagation by more accurately resolving pressure gradients, leading to a reduction of the maximum velocity error to $4.6 \times 10^{-3}$, a significantly smoother error field, and a relative velocity $L^2$ error of $1.2 \times 10^{-3}$, thereby ensuring improved global consistency across both velocity and pressure fields.

From a spatial discretization (see Figures~\ref{fig:caseVconvwith} and     \ref{fig:caseVconvwithout}) perspective, the baseline employs a fixed uniform sampling of collocation points, which fails to capture localized gradients near the inlet and wall regions due to their higher gradient. The AQPINN instead performs residual-driven adaptive re-meshing, dynamically reallocating points toward high-error regions, particularly near the inlet boundary layer ($x \to 0$) and the core flow region around $x \approx 2.0$. This adaptive refinement improves resolution of localized stiffness without increasing network complexity or parameter count, highlighting the effectiveness of residual-informed point placement in multi-field fluid dynamics.

\subsubsection{\label{sec:NS-TC}
Navier-Stokes equations - Taylor-Couette flow}

Finally, as a last case, we consider the steady, incompressible Navier–Stokes equations for viscous flow between two concentric, infinitely long rotating cylinders in polar coordinates over the annular domain $\Omega={(r,\theta),|,r\in[R_i,R_o],,\theta\in[0,2\pi]}$. Under axisymmetry, the velocity field $\mathbf{u}=(u_r,u_\theta)$ satisfies the radial and azimuthal momentum equations coupled with continuity:
\begin{align}
u_r \frac{\partial u_r}{\partial r} - \frac{u_\theta^2}{r} 
&= -\frac{1}{\rho}\frac{\partial p}{\partial r} 
+ \nu \left( \frac{\partial^2 u_r}{\partial r^2} + \frac{1}{r}\frac{\partial u_r}{\partial r} - \frac{u_r}{r^2} \right), \label{eq:momentum_r} \\
u_r \frac{\partial u_\theta}{\partial r} + \frac{u_r u_\theta}{r} 
&= \nu \left( \frac{\partial^2 u_\theta}{\partial r^2} + \frac{1}{r}\frac{\partial u_\theta}{\partial r} - \frac{u_\theta}{r^2} \right), \label{eq:momentum_theta} \\
\frac{\partial u_r}{\partial r} + \frac{u_r}{r} &= 0. \label{eq:continuity}
\end{align}

No-slip boundary conditions are imposed at $R_i=1$ and $R_o=2$, with $u_\theta(R_i)=\Omega_i R_i$ and $u_\theta(R_o)=\Omega_o R_o$, where $\Omega_i=1$ and $\Omega_o=0$. This yields the classical analytical Couette solution:
\begin{equation}
u_\theta(r)=Ar+\frac{B}{r}, \quad u_r(r)=0,
\end{equation}
with
\begin{equation}
A=\frac{\Omega_o R_o^2-\Omega_i R_i^2}{R_o^2-R_i^2}, \quad B=\frac{(\Omega_i-\Omega_o)R_i^2R_o^2}{R_o^2-R_i^2},
\end{equation}
and pressure obtained from $p_r=\rho,u_\theta^2/r$.

The model comprises a classical projection MLP  with 2 Layers with 32 neurons with SiLU activations, followed by a 4-qubit variational quantum circuit (VQC) with depth 4 Layers, and a classical decoding MLP with 2 Layers  with 32 neurons with SiLU that reconstructs the physical state vector $\hat{\mathbf{y}}(r,\theta) = [u_r, u_\theta, p]^T$.

Adaptive training is performed using a residual-driven collocation refinement strategy initialized with 200 interior points sampled via Latin hypercube sampling (LHS), and dynamically expanded up to a maximum of 600 points based on local error indicators. A minimum Euclidean spacing of 0.02 is enforced to prevent point clustering and maintain spatial coverage quality during refinement. The optimization monitors five coupled loss components corresponding to the radial and azimuthal momentum equations. A moving-average exponential smoothing is applied to the residual field to stabilize spatiotemporal error tracking and improve robustness of refinement decisions. 

An examination of the continuous field topologies shows that both the baseline QPINN and the AQPINN accurately resolve the azimuthal velocity field $u_\theta$ in the Taylor--Couette configuration. The cross-sectional profiles at $\theta=\pi$ exhibit excellent agreement with the logarithmic-like analytical decay from $u_\theta=1$ at the inner cylinder to $u_\theta=0$ at the outer wall. The baseline achieves RMSE $3.299\times10^{-4}$ and $R^2=0.99$, while the AQPINN improves this to RMSE $3.122\times10^{-5}$ and $R^2\approx1.0$, indicating near-exact recovery of the macroscopic flow structure. However, localized error analysis reveals persistent boundary-layer artifacts in the baseline, with peak absolute error $6.427\times10^{-4}$ concentrated near the outer quadrant, which is significantly reduced by the AQPINN to $9.501\times10^{-5}$ ($85.2\%$ reduction). In the radial component $u_r$, the baseline exhibits structured residual rings with maxima of $4.481\times10^{-8}$, whereas the AQPINN suppresses these coherent artifacts into dispersed low-amplitude fluctuations ($3.883\times10^{-8}$), improving mass-conservation fidelity.

These improvements are reflected in the multi-objective optimization dynamics, where the standard QPINN employs fixed weighting, leading to stagnation of the physics residual near $10^{-5}$. In contrast, again, the AQPINN employs adaptive loss reweighting that maintains gradient diversity and avoids multi-objective collapse, driving further down to $\sim10^{-6}$ and achieving a best training loss of $1.0\times10^{-6}$ versus $1.3\times10^{-5}$ for the baseline. This is complemented by a residual-driven adaptive collocation strategy, where uniformly distributed nodes in the baseline fail to resolve localized stiffness, whereas the AQPINN dynamically re-allocates collocation points toward high-residual regions near the outer boundary layer and shear zones, thereby enhancing local resolution without increasing model complexity.

\subsection{\label{sec:Conclusion}Conclusion}

In this work, we have systematically investigated the performance and structural limitations of standard QPINNs and introduced a comprehensive evaluation of the AQPINN framework. By evaluating both methodologies across six distinct numerical benchmarks-spanning canonical quantum harmonic oscillators, highly non-linear fluid dynamics, and complex curvilinear boundary flows-we have demonstrated the critical pathologies that lead to optimization stagnation in traditional fixed-hyperparameter PINN formulations, and showcase how dynamic adaptation effectively mitigates these challenges.

Our comparative findings yield several major insights into quantum physics-informed deep learning: \textbf{mitigation of Gradient Pathologies:} standard QPINNs utilizing static objective weights ($\alpha_i$) are highly susceptible to gradient stiffness, often settling into unphysical trivial states or experiencing premature optimization plateaus. Conversely, the soft-adaptive attention mechanism embedded within the AQPINN framework dynamically re-balances the competing gradients of governing differential equations, initial conditions, and boundary terms. This enables the network to escape local minima, achieving error reductions at least twice. \textbf{automated geometric re-Meshing:} Static, uniform collocation point distributions struggle to capture high-gradient regions, moving fronts, or multi-field dependencies. By leveraging local residual history tracking, the AQPINN automatically aggregates and clusters training nodes directly within localized spatial and temporal regions exhibiting extreme mathematical stiffness. This strategy successfully captured moving shock profiles in the non-linear viscous Burgers' equation and localized boundary layer effects under coordinate transformations without inflating the overall model parameter size or necessitating massive training sets. \textbf{Robustness in Multi-Field and Coupled Configurations:} In systems characterized by severe scale discrepancies between coupled field variables, such as the axisymmetric Hagen-Poiseuille pipe flow and the annular Taylor-Couette flow, standard models show a stark divergence in predictive fidelity across variables-frequently resolving primary fields while inducing unphysical topologies in secondary conservation metrics. The AQPINN architecture enforces strict mathematical and physical consistency across coupled domains, capturing delicate pressure gradients and satisfying the divergence-free continuity constraint to a high level of numerical precision.

Ultimately, the results across these six benchmarks demonstrate that self-adaptive attention mechanics and history-driven node adaptation enhance performance in complex engineering domains.
Future research will focus on extending the AQPINN architecture to higher-dimensional problem domains, scaling the adaptive re-meshing scheme to fully three-dimensional fluid simulations, and investigating the limits of coupled adaptive systems within stiff partial differential equation loss landscapes.

\subsection*{\label{sec:Appen}Appendix}
In this appendix, we analyze the importance of each adaptation mechanism: self-attention and adaptive collocation points. We selected two complex problems to independently evaluate the contribution of each mechanism. We conclude that both techniques are effective on their own; however, when combined, they lead to significantly enhanced performance. The results are presented below.
\subsubsection*{Burgers' equation analysis}
The empirical evaluation of the QPINN variants on the Burgers' equation demonstrates the complementary contributions of the proposed enhancements. The baseline model (QPINN) struggles to capture the strong nonlinearities and shock-like gradients, yielding an RMSE of $1.917 \times 10^{-1}$ and an $R^2$ of $0.89$. Introducing point-adaptive collocation (QPINN-collocation) improves accuracy by concentrating collocation points in high-gradient regions, reducing the RMSE to $1.2 \times 10^{-1}$ and increasing $R^2$ to $0.95$.

Replacing spatial adaptation with the self-attention mechanism (QPINN-self-attention) provides a greater improvement, reducing RMSE to $9.306 \times 10^{-2}$ and increasing $R^2$ to $0.97$. By dynamically weighting spatio-temporal features and loss contributions, self-attention enables the network to better capture long-range dependencies and nonlinear interactions.

The combined architecture (QPINN-FULL) achieves the best performance across all metrics, highlighting the synergy between adaptive collocation and self-attention. It achieves the lowest RMSE ($6.925 \times 10^{-2}$), highest $R^2$ ($0.986$), lowest maximum error ($6.479 \times 10^{-1}$), and lowest training loss ($2.360 \times 10^{-5}$), corresponding to an overall RMSE reduction of approximately 64\% relative to the baseline. These results demonstrate that integrating spatial and feature-based adaptivity yields a robust and accurate framework for solving highly nonlinear partial differential equations.

\begin{table}[htbp]
\centering
\caption{Comparison of Error Metrics across QPINN Architecture Cases}
\label{tab:error_metrics_comparison2}
\begin{tabular}{lcccc}
\toprule
\textbf{Error Metric} & \textbf{QPINN} & \textbf{QPINN-collocation} & \textbf{QPINN-self-attention} & \textbf{AQPINN} \\RMSE            & $1.917 \times 10^{-1}$  & $1.203 \times 10^{-1}$  & $9.306 \times 10^{-2}$  & $6.925 \times 10^{-2}$ \\
MAE             & $3.441 \times 10^{-2}$  & $2.464 \times 10^{-2}$  & $1.757 \times 10^{-2}$  & $1.351 \times 10^{-2}$ \\
Max error       & $1.950 \times 10^{0\phantom{-}}$ & $1.069 \times 10^{0\phantom{-}}$ & $9.114 \times 10^{-1}$ & $6.479 \times 10^{-1}$ \\
Rel $L_2$       & $3.230 \times 10^{-1}$  & $2.027 \times 10^{-1}$  & $1.568 \times 10^{-1}$  & $1.167 \times 10^{-1}$ \\
Rel $L_\infty$  & $1.952 \times 10^{0\phantom{-}}$ & $1.069 \times 10^{0\phantom{-}}$ & $9.121 \times 10^{-1}$ & $6.484 \times 10^{-1}$ \\
$R^2$           & 0.895669                & 0.958901                & 0.975411                & 0.986384 \\
Best loss       & $3.635 \times 10^{-4}$  & $3.244 \times 10^{-5}$  & $4.016 \times 10^{-5}$  & $2.360 \times 10^{-5}$ \\
\end{tabular}
\end{table}

\subsubsection*{Taylor-Couette flow analysis}

The comparative analysis of QPINN architectures on the Taylor--Couette flow benchmark focuses on the azimuthal velocity component ($u_\theta$). The baseline model (QPINN) gets an RMSE of $3.299 \times 10^{-4}$, an MAE of $3.067 \times 10^{-4}$, and an $R^2$ of $0.99$. Nevertheless, residual errors remain in localized regions, as reflected by a maximum error of $7.268 \times 10^{-4}$ and a relative error of $L_\infty$ of $6.427 \times 10^{-4}$.

Introducing point-adaptive collocation (QPINN-collocation) improves accuracy by refining sampling in high-shear regions between the rotating cylinders. This reduces the RMSE to $1.441 \times 10^{-4}$ and the MAE to $1.173 \times 10^{-4}$, while also lowering the maximum error to $4.407 \times 10^{-4}$ and achieving a best loss of $1.710 \times 10^{-6}$. In contrast, the self-attention-only variant (QPINN-self-attention) provides a more moderate improvement, reaching an RMSE of $2.450 \times 10^{-4}$ and MAE of $1.785 \times 10^{-4}$, but with a higher maximum error of $6.427 \times 10^{-4}$, indicating that spatial resolution is more critical than feature reweighting for this smooth flow regime.

The combined architecture (QPINN-FULL) delivers the strongest performance across all metrics, confirming a clear synergy between adaptive collocation and self-attention. It achieves a RMSE of $3.122 \times 10^{-5}$, MAE of $2.789 \times 10^{-5}$, and a maximum error reduced to $9.501 \times 10^{-5}$, corresponding to an approximately 90\% reduction in RMSE relative to the baseline. The model also attains an $R^2$ of $1.0$ and the lowest best loss of $1.093 \times 10^{-6}$. These results demonstrate that combining spatial adaptivity with dynamic feature weighting enables a great reconstruction even in smooth, geometrically constrained fluid dynamics problems.

\begin{table}[htbp]
\centering
\caption{Comparison of Error Metrics across QPINN Architecture Cases}
\label{tab:metrics_comparison1}
\begin{tabular}{lcccc}
\hline \hline
\textbf{Error Metric} & \textbf{QPINN} & \textbf{QPINN-collocation} & \textbf{QPINN-self-attention} & \textbf{AQPINN} \\
\multicolumn{5}{l}{\textbf{Error metrics ($u_\theta$)}} \\
RMSE & $3.299 \times 10^{-4}$ & $1.441 \times 10^{-4}$ & $2.450 \times 10^{-4}$ & $3.122 \times 10^{-5}$ \\
MAE & $3.067 \times 10^{-4}$ & $1.173 \times 10^{-4}$ & $1.785 \times 10^{-4}$ & $2.789 \times 10^{-5}$ \\
Max error & $7.268 \times 10^{-4}$ & $4.407 \times 10^{-4}$ & $6.427 \times 10^{-4}$ & $9.501 \times 10^{-5}$ \\
Rel $L_2$ & $6.418 \times 10^{-4}$ & $2.693 \times 10^{-4}$ & $4.579 \times 10^{-4}$ & $6.075 \times 10^{-5}$ \\
Rel $L_\infty$ & $6.427 \times 10^{-4}$ & $4.407 \times 10^{-4}$ & $7.268 \times 10^{-4}$ & $9.501 \times 10^{-5}$ \\
$R^2$ & 0.999 & 1.0 & 0.999 & 1.0 \\
Best loss & $1.387 \times 10^{-5}$ & $1.710 \times 10^{-6}$ & $5.234 \times 10^{-6}$ & $1.093 \times 10^{-6}$ \\
\end{tabular}
\end{table}

\subsection*{\label{sec:Ack}Acknowledgement}

This work was supported by the Serrapilheira Institute (grant number Serra – R-2111-39718).
%---------------------------------------------------------------
\subsection*{\label{sec:Ethics} AI-assisted Ethics}

AI-assisted Content Development: the authors used artificial intelligence tools, including ChatGPT and Grammarly, to support language refinement and improve readability. The scientific content, interpretations, and final text were fully reviewed and validated by the authors, who assume full responsibility for the manuscript.

\newpage
\subsection{\label{sec:Ref}References}

\bibliographystyle{unsrtnat}
\bibliography{apssamp}% Produces the bibliography via BibTeX.

\end{document}